\documentclass[10pt,twocolumn,letterpaper]{article}

\usepackage[pagenumbers]{cvpr}       
\usepackage[accsupp]{axessibility}
\usepackage{graphicx}
\usepackage{amsmath}
\usepackage{amssymb}
\usepackage{booktabs}
\usepackage{bbm}
\usepackage{subfiles}

\usepackage{algorithm}
\usepackage{algpseudocode}
\usepackage[pagebackref,breaklinks,colorlinks]{hyperref}

\usepackage[capitalize]{cleveref}
\crefname{section}{Sec.}{Secs.}
\Crefname{section}{Section}{Sections}
\Crefname{table}{Table}{Tables}
\crefname{table}{Tab.}{Tabs.}

\def\cvprPaperID{5948} % *** Enter the CVPR Paper ID here
\def\confName{CVPR}
\def\confYear{2022}

\begin{document}

\title{CellTypeGraph: A New Geometric Computer Vision Benchmark}

\author{
Lorenzo Cerrone\textsuperscript{\rm 1}\\
%{\tt\small HCI Heidelberg University}\\
\and 
Athul Vijayan\textsuperscript{\rm 3}\\
%Department of Comparative Developmental and Genetics\\
%MPI for Plant Breeding Research, Cologne, Germany\\
%{\tt\small MPI Cologne Germany}\\
\and
Tejasvinee Mody\textsuperscript{\rm 2}\\
%Department of Comparative Developmental and Genetics\\
%MPI for Plant Breeding Research, Cologne, Germany\\
%{\tt\small TUM Germany}\\
%{\tt\small avijayan@mpipz.mpg.de}
\and
Kay Schneitz\textsuperscript{\rm 2}\\
%School of Life Sciences\\
%{\tt\small TUM Germany}\\
%{\tt\small kay.schneitz@tum.de}
\and
Fred A. Hamprecht\textsuperscript{\rm 1}\\
%{\tt\small Heidelberg University Germany}\\
%{\tt\small fred.hamprecht@iwr.uni-heidelberg.de}
%{\tt\small [lorenzo.cerrone, fred.hamprecht]@iwr.uni-heidelberg.de avijayan@mpipz.mpg.de kay.schneitz@tum.de}
\\
\textsuperscript{\rm 1}HCI, Heidelberg University, Germany\\
\textsuperscript{\rm 2}School of Life, Technical University of Munich, Germany\\
\textsuperscript{\rm 3}Max Planck Institute for Plant Breeding Research, Germany\\
{\tt\small lorenzo.cerrone@iwr.uni-heidelberg.de}\\
}
\maketitle

%%%%%%%%% ABSTRACT
\begin{abstract}
Classifying all cells in an organ is a relevant and difficult problem from plant developmental biology. We here abstract the problem into a new benchmark for node classification in a geo-referenced graph. Solving it requires learning the spatial layout of the organ including symmetries. 
To allow the convenient testing of new geometrical learning methods, the benchmark of \textit{Arabidopsis thaliana} ovules is made available as a  PyTorch data loader, along with a large number of precomputed features. Finally, we benchmark eight recent graph neural network architectures, finding that \textit{DeeperGCN} currently works best on this problem.
\end{abstract}

%%%%%%%%% BODY TEXT
\section{Introduction}
\label{sec:intro}
Understanding morphogenesis, the generation of form, remains a major challenge in biology. It requires a detailed quantitative description of the molecular and cellular processes of the underlying mechanism. 3D digital organs with cellular spatial resolution promise to help decipher the morphogenesis of complex organs in plants \cite{montenegro2015digital, willis2016cell, refahi2021multiscale, hong2018heterogeneity}. They can be generated by 3D microscopic imaging followed by cell instance segmentation and tissue annotation.
Plant organs lend themselves to this approach as they exhibit a relatively well-structured, layered organization and the tissues can be identified based on their positions and morphology. Thus, plant developmental biology is profiting from a growing body of 3D digital organs. However, automatic annotation \cite{montenegro2015digital, schmidt2014irocs, montenegro20193dcellatlas} of different tissue types within an organ remains a major problem in the field, particularly for plant organs of complex cellular architecture and shape such as ovules of \textit{Arabidopsis thaliana} \cite{vijayan2021digital}.
%Different methods exist for automated tissue annotations in simple and layered tissues of plant organs \cite{montenegro2015digital, schmidt2014irocs, montenegro20193dcellatlas}, but these methods fail to classify cells of complex shaped organs such as ovules of Arabidopsis thaliana \cite{vijayan2021digital}.
%\cite{ronneberger2015u, cciccek20163d}

In medical image analysis, the analogous problem of segmenting whole-body scans has been addressed successfully using 3D encoder-decoder CNN architectures \cite{weston2019automated, lee2017pixel}. In the images of interest to plant morphogenesis, this straightforward approach does not work as well: here, the semantic classes lack distinctive local and texture features helpful for convolution based architectures \cite{geirhos2018imagenet}. Instead, the task requires very long-range spatial awareness and good geometrical reasoning. In response, here we cast the problem as a node classification task. To succeed, any approach requires informative input features, and indeed we show that a cell-adjacency graph with cell-level features is a powerful representation.

That said, our primary intent is twofold: first, to create a testing ground for machine learning on highly structured input.
Indeed, it uniquely combines node classification typical of popular datasets \cite{mccallum2000automating, giles1998citeseer, sen2008collective} with the requirement to generalize across geo-referenced graphs like in Quantum Chemistry \cite{rupp, blum, wu2018moleculenet, gomez2018automatic}.
Our second intent is, by allowing computer vision and machine learning scientists to work on this problem without having to deal with data-processing issues that actually take the most time in many applied projects, to channel the creativity and resourcefulness of our community to help solve a fascinating biological problem. 

As a starting point, we present extensive experiments evaluating the performance of state-of-the-art and popular models.

In summary, we make the following contributions:
\begin{enumerate}
    \item We propose a new benchmark for node classification in a geo-referenced graph. We release the benchmark dataset together with a ready-to-use PyTorch \cite{paszke2017automatic} data loader. The source-code and usage instructions are available at \url{https://github.com/hci-unihd/celltype-graph-benchmark}.
    \item We run comparative experiments using state-of-the-art graph neural networks and offer a study of feature relevance.
    \item We provide an extensive set of precomputed features and an additional set of ground truth labels.
\end{enumerate}

%-------------------------------------------------------------------------
\subsection{Related work}

Arguably the most popular type of dataset for node predictions is the family of citation datasets, such as Cora \cite{mccallum2000automating}, CiteSeer \cite{giles1998citeseer} or PubMed \cite{sen2008collective}. 
Although the task is nominally identical, the learning task is here transductive, and the underlying topologies are extremely different. In our CellTypeGraph, the task is inductive, and nodes are geo-referenced with a limited variance in node degrees and no shortcuts between distant nodes. This is not typically the case in citation datasets.
More closely related are other natural science datasets such as QM9 \cite{rupp, blum} or ZINC \cite{gomez2018automatic}, MoleculeNet \cite{wu2018moleculenet}. Nodes here represent atoms and are geo-referenced. Nevertheless, they are, for the most part, interested in regression and classification of global graph properties.

The Open Graph Benchmark \cite{hu2020open} contains a vast collection of datasets at different scales and also releases easy to use data loader and evaluation tools. As pointed out in \cite{hu2020open}, on many datasets the limited size, lack of agreed-upon train/val/test splits, and use of different metrics make it hard to measure progress in the field. In CellTypeGraph, we also release all tools required to ensure simple and reproducible experimentation. Moreover, we propose cross-validation \cite{friedman2017elements} rather than a simple train/val/test split to further reducing sampling biases.

Motivated by the success of convolution neural networks, several graph neural network architectures attempt to generalize the same concept to non-grid structured data \cite{kipf2016semi, defferrard2016convolutional, he2021bernnet}. Those architectures and many more \cite{velickovic2018graph, hamilton2017inductive, xu2018powerful, pmlr-v97-wu19e, chen2020simple, li2019deepgcns, li2020deepergcn, brody2021attentive} can be modeled as message passing \cite{gilmer2017neural}, where messages built at the node level can be shared over the local neighborhood.
In case of geo-referenced or spatial graphs, the node positions can be used to model the messages aggregations over the graph adjacency \cite{danel2020spatial, de2020gauge}. Alternatively, others \cite{schutt2017schnet, unke2019physnet}, have obtained competitive results on molecular tasks by using only node positions and ignoring the graph adjacency.

\section{CellTypeGraph benchmark}

\begin{figure*}
  \centering
  \includegraphics[width=1\linewidth]{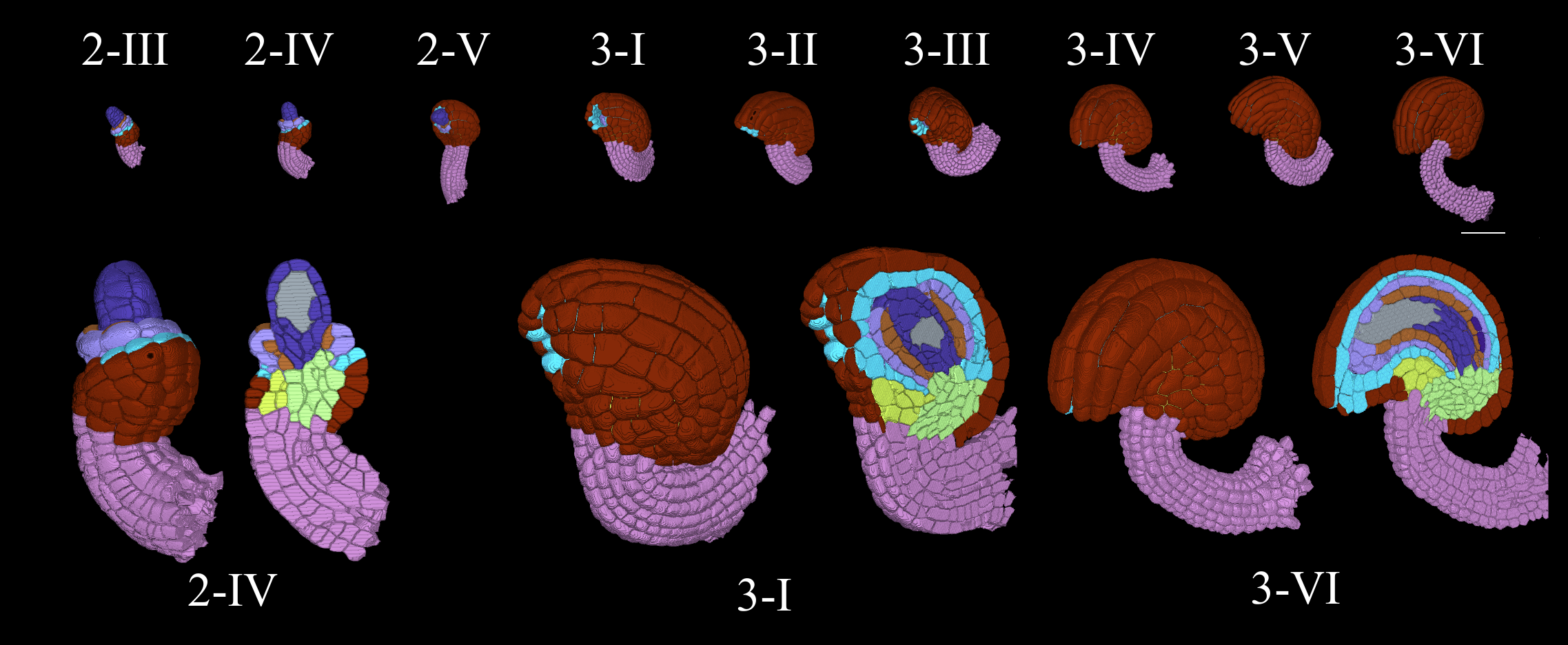}
  \caption{3D surface view of a small subset of specimens from the CellTypeGraph Benchmark. Different developmental stages are indicated. From left to right the tissue complexity increases with organ growth. Scale bar $50 \mu m$.
 Bottom: three stages are represented with their 3D view and a 2D section displaying the internal tissue architecture. Colors show ground truth cell types.}
  \label{fig:main}
\end{figure*}

\subsection{Overview}
We introduce the CellTypeGraph Benchmark, aiming to give a valuable tool to the geometric learning community. To do so, we distilled a publicly available biological dataset \cite{vijayan2021digital} into a ready-to-use machine learning benchmark.
In particular, we would like to focus the community's attention on two primary objectives:
\begin{itemize}
    \item Finding better graph neural network architectures and methodologies. In this case, we encourage using our pre-computed features, thus allowing direct comparison with our baseline, see \cref{sec:baseline}.
    \item Finding more expressive features or, alternatively, replace hand crafted features by end-to-end learning. We will expand on this point in \cref{subsec:basic-feat}.
\end{itemize} 
Progressing in either of the two challenges is an equally valid contribution to achieving the final goal of improving automated tools for plant-biology.

The statistics of the dataset are summarized in \cref{tab:summary}.

\subsection{Raw data}
The benchmark dataset is obtained from 3D confocal microscopic images of Arabidopsis ovule \cite{vijayan2021digital}. Ovules are the female gametophyte in higher plants that eventually gives rise to seeds upon successful fertilization.

The 3D digital ovule atlas is a composite of raw cell boundary images and their respective cell segmentation which are further tissue annotated. \cref{fig:main} shows a sample ovule from each developmental stage. Ovules at early stages (2-III to 2-V) protrude from the placental surface as a simple three layered dome like structure. Ovules form a complex 3D tissue structure during integument initiation and outgrowth. Four layers of integument tissues surround the core of the organ (cell labels \textit{L1} to \textit{L4}). Integument tissues undergo growth conflicts that result in their final shape with a hood-like outer structure at maturity. Inner tissue layers of integuments are arranged as a bent tube like shape.
Overall, the organ is moulded by complex arrangements of cells in 3D that partly follow a stratified arrangement of cells in the medial and upper half of the organ and it also forms a partial radial and bilateral symmetry.

The ground truth cell type labels for the benchmark dataset were obtained semi-automatically by first using a modified detect layer stack process in MorphographX \cite{strauss2021morphographx}; and then proofreading the results from these predictions manually. Essentially, the layered cells in the upper medial half of the organ were labeled by the cell layer detection method and then manually corrected, while the rest of the cells from the organ were labeled manually.
%For manual corrections, different tissues were manually identified based on the cellular connectivity and position in space, they were then separated from the organ mesh to perform tissue annotations in MorphographX. 
This required extensive human input, totaling in the order of 60 human hours for the dataset in this study.

Tissue annotations allowed in extensive biological analysis of the cellular basis of growth in different tissues and at different developmental stages \cite{vijayan2021digital}. 
%Overall, a robust method for fully automated tissue annotation of Arabidopsis ovule is essential as the organ is now extensively studied with a higher number of samples being analyzed and that the current methods require extensive human corrections.

\paragraph{Pre-processing:}
The segmentation images as published in \cite{vijayan2021digital} have been further manually curated. We have ensured that for each specimen, cells form a single connected component. In case of multiple specimen imaged together, they have been split in separate stacks. Lastly, cell type annotations have been added for missing cells.
Although we did our best to hand curate the data, minor imperfections might still be present due to the variability of the organs.
Since the \textit{L5} cell type is rarely present in the later stages, we slightly simplified the benchmark task by merging the cell types \textit{L4} and \textit{L5}.

\begin{table}
  \centering
  \begin{tabular}{@{}lr@{}}
    \toprule
    Number of specimens & 84 \\
    Number of developmental stages & 9 \\
    Total number of cells/nodes & 95757\\
    Total number of edges & 632443 \\
    Average number of nodes per specimen & 1140 \\
    Average number of edges per specimen & 7529 \\
    Number of semantic classes & 9 \\
    Number of node features & 78\\
    Number of edge features & 11\\
    \bottomrule
  \end{tabular}
  \caption{Salient statistics of the CellTypeGraph benchmark.}
  \label{tab:summary}
\end{table}

\begin{figure}
    \centering
    \includegraphics[width=1\linewidth]{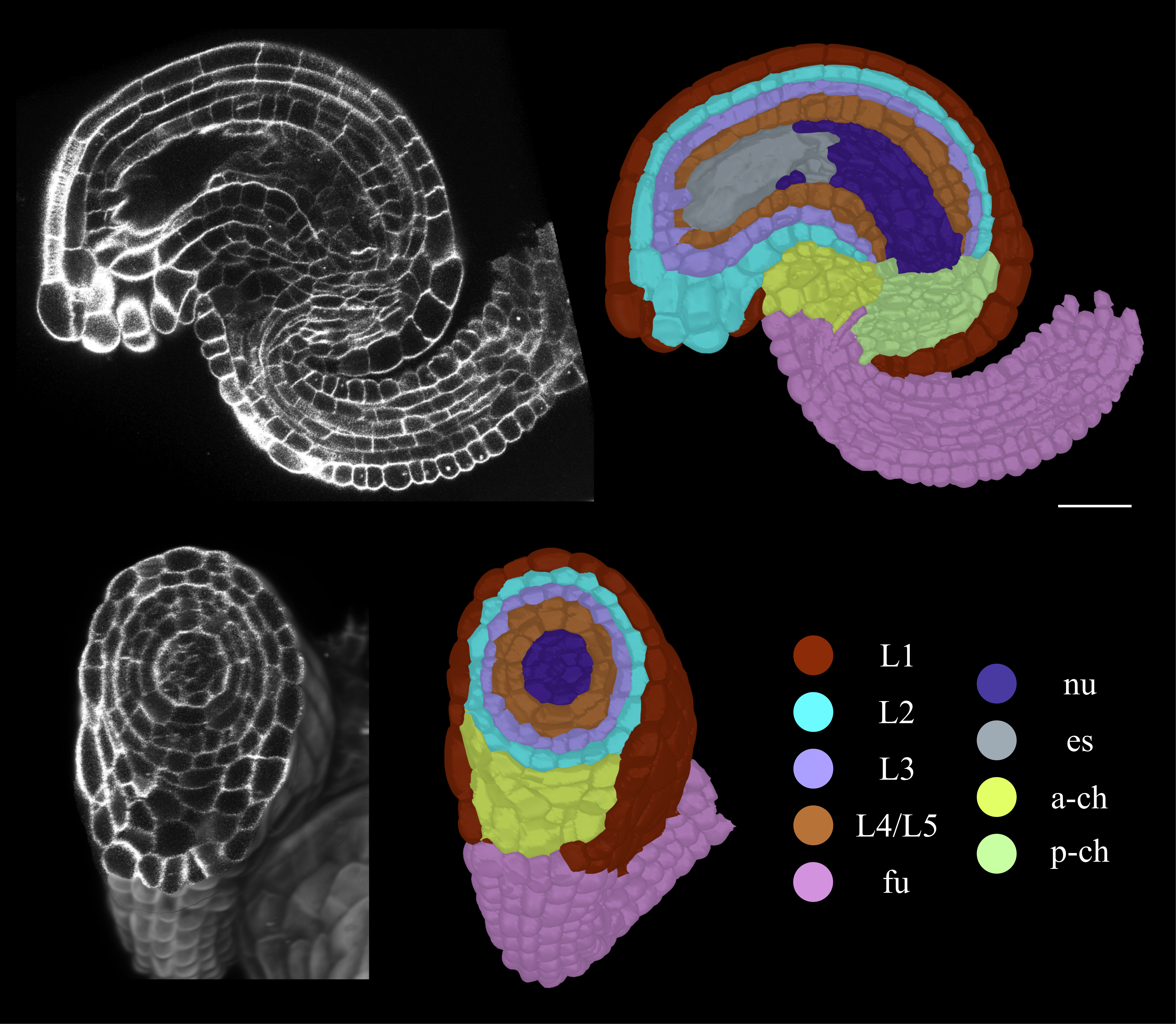}
  \caption{2D section view of a mature \textit{Arabidopsis} ovule displaying the raw cell boundary image and the respective CellTypeGraph ground truth labels manually annotated for the benchmark dataset. Different colors indicate different tissue labels annotated to the 3D instance cell segmentation. Abbreviations \textit{es}: embryo sac, \textit{nu}: nucellus, \textit{L1}: outer layer of outer integument, \textit{L2}: inner layer of outer integument, \textit{L3}: outer layer of inner integument, \textit{L4}/\textit{L5}: inner layer of inner integument, \textit{fu}: funiculus, \textit{a-ch}: anterior chalaza, \textit{p-ch}: posterior chalaza. Scale bar $20 \mu m$.}
  \label{fig:labels-legend}
\end{figure}

\subsection{Evaluation}
\label{subsec:evaluation}
\paragraph{Metrics:} The ovules cell types are imbalanced. This not only impacts the learning procedure but also requires attention when discussing the results quantitatively.
An effective way to account for imbalance is to evaluate the model performance for each class independently and then report as final score the average of the single class results.
In this work, we use two metrics, the simple global \textit{top-1 accuracy} and the \textit{class-average accuracy}.

In order to further stabilize the \textit{class-average accuracy}, we ignored cell label $7$, see \cref{fig:labels-legend}.
Cell label $7$ represents the ``embryo sac''. This tissue is not a cell in itself but is an ensemble of non-segmentable cells in the inner part of later-stage ovules. This region is unique and distinct, being the largest and highest degree node in the cell graph. Nevertheless, we kept the ``embryo sac'' in the benchmark for training purposes because of its crucial role in graph connectivity.

Moreover, if any cell type is not present in a specific specimen, that cell type will not be counted in the \textit{class-average accuracy}, i.e.,
\begin{equation*}
    \textit{class-average accuracy} = \frac{1}{N_s}\sum_{s=1}^{N_s} \frac{\sum_c a_c \cdot \mathbbm{1}_{s, c}}{\sum_c \mathbbm{1}_{s, c}}
\end{equation*}
where $N_s$ is the number of specimens, $a_c$ is the one-vs-all class accuracy, $\mathbbm{1}$ is an indicator function valued $1$ if the class $c$ is present in the specimen $s$, 0 otherwise. This is particularly relevant in the early stages of development.

We release evaluation code for all the metrics introduced above bundled with our benchmark.

\paragraph{Experts consensus}
The cell types are identified based on patterns of gene expression, cell position in space, and context in terms of tissue layers. The consensus is strong for layered cell types (\textit{L1} to \textit{L4}); here, segmentation quality is the dominant source of ambiguity. But variability can arise in regions where the boundaries of the tissue groups are hard to determine from the cell position itself; for example, the boundary between \textit{fu}, \textit{ch}, and \textit{nu}.
In order to assess this variability quantitatively and assess a reference performance for an expert biologist, we produced an independent second set of labels.
%In order to set a reference performance for an expert biologist on our benchmark task, we evaluated our metrics on our independent set of labels. 
The overall results are reported in \cref{tab:node_class_eval}, while an additional breakdown of expert performance at different developmental stages and for different classes is reported in \cref{suppl:experts}.

\paragraph{Training and evaluation splits:} We release our CellTypeGraph data loader in two different modes: a standardized train-validation-test split, and a five fold cross-validation split.

Although train-validation-test splits are standard practice in popular machine learning benchmarks, this approach makes evaluations susceptible to sampling biases \cite{shchur2018pitfalls}. In our CellTypeGraph, we have high variability between specimens both intra-stage and inter-stage, see \cref{fig:main}.
This, combined with the relatively small number of specimens, makes the splits highly susceptible to sampling bias.
We address the stage variability by sampling the same number of specimen from each stage.

However, to further unbias our experiments, we preferred to focus on a five-fold cross-validation approach. Cross-validation mitigates the noise coming from the random sampling at the expense of a more costly training. All presented experiments have been evaluated using cross-validation.

\section{Features}
This section details the node and edge features included in the CellTypeGraph benchmark. However, before motivating their choice, it is necessary to discuss the reference system used to express position and orientation dependent features. 
\label{sec:features}
\subsection{Reference systems}
\paragraph{Global reference system:}
\label{subsec:grs}
Ovule images acquired by the microscope are not always oriented in the same direction. This inconsistency in specimen orientation could cause poor generalization of trained models. Possible solutions are to systematically use rotation and translation equivariant methods or to fix a landmark-based global reference system. The first solution is more general and shows great potential in terms of generalization and parameter efficiency \cite{finzi2020generalizing, keriven2019universal, satorras2021n}.
Equivaraint methods have a clear advatage when no standardized orientation can be formed, as in general chemical problems. In our case, an unambiguous standard pose can be defined and so for simplicity we opt for a landmark-based approach.

\begin{figure}
    \centering
    \includegraphics[width=1\linewidth]{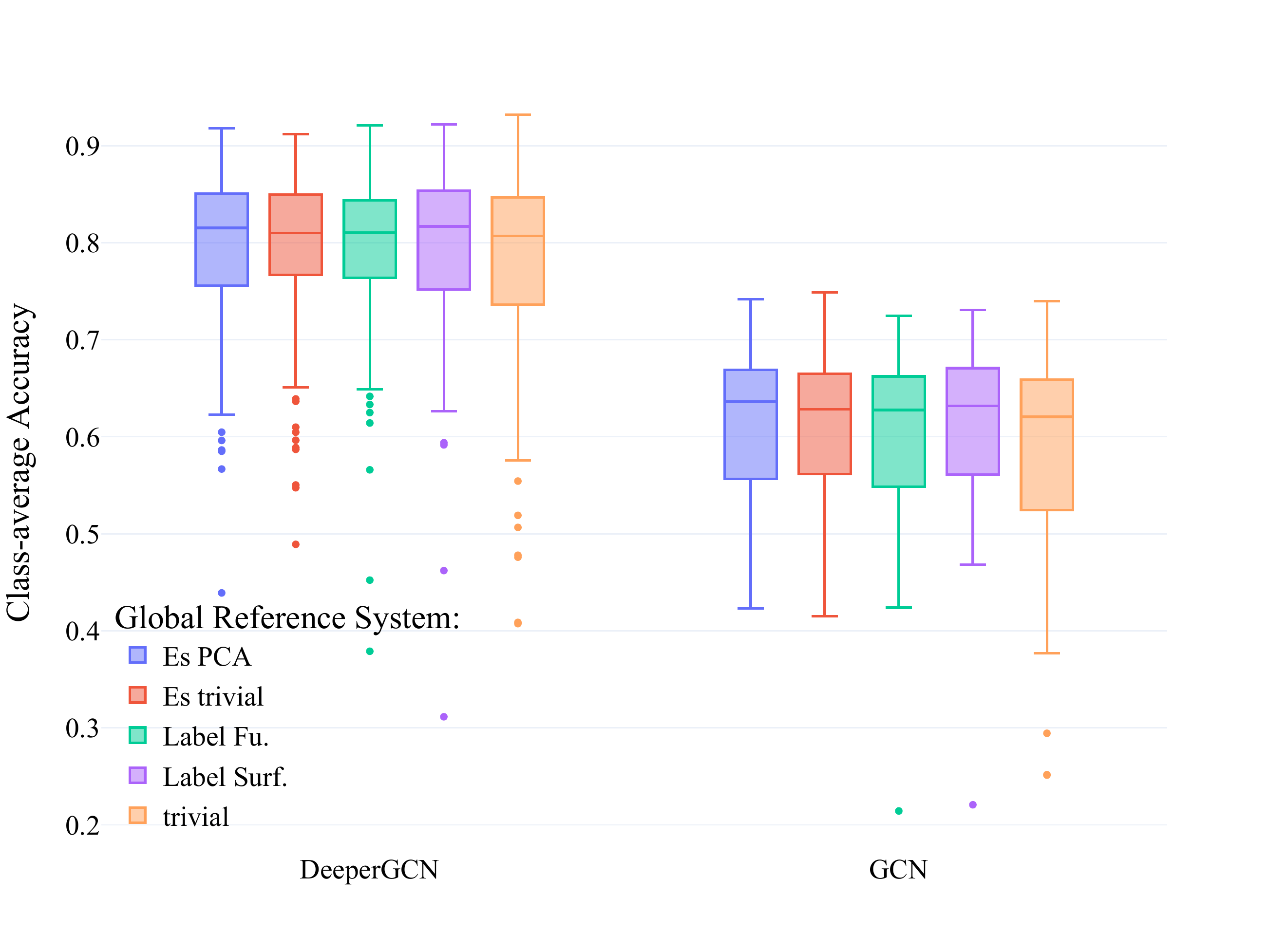}
  \caption{All reference systems are equally good. We trained both \textit{GCN} and \textit{DeeperGCN} on five distinct reference systems. There is no statistically significant difference in using either reference system. The only difference we could observe was in the spread of outliers.}
  \label{fig:grs_stability}
\end{figure}

More specifically, we evaluate four landmark-based approaches, both supervised (\textit{Label Surf}, \textit{Label Fu}) and unsupervised (\textit{ES trivial}, \textit{ES PCA}). Detailed descriptions can be found in \cref{suppl:sec1}.
All coordinate systems yield similar accuracy, see \cref{additiona_exp}. Nevertheless, the local reference \textit{Label Surf} is slightly more consistent and less prone to outliers compared with the others, see \cref{fig:grs_stability}. Hence \textit{Label Surf} was used as default orientation system for all of our experiments. 

\paragraph{Local reference system:}
Ovule cells have no distinctive anisotropy that can be used to define a local orientation.
However, preferential direction of growth and divisions along the central curved axis of the organ result in filar arrangement of cells within integument tissue layers (\textit{L1} to \textit{L4}). This regular filar arrangement of cells can be used to identify integument tissues and separate them from one another. Integument tissues divide anticlinally to maintain their sheet like structure. Cell divisions happen mainly along the transverse anticlinal direction allowing the tissue to grow in length. Division planes can be further mapped to faces of cell wall extracted from a 3D instance cell segmentation. This includes longitudinal anticlinal wall, transverse anticlinal wall and periclinal wall. In order to exploit this prior, we construct a heuristic algorithm to identify transverse anticlinal walls and define a \textit{growth axis} based on their connectivity.

\begin{figure}
  \centering
  \includegraphics[width=1\linewidth]{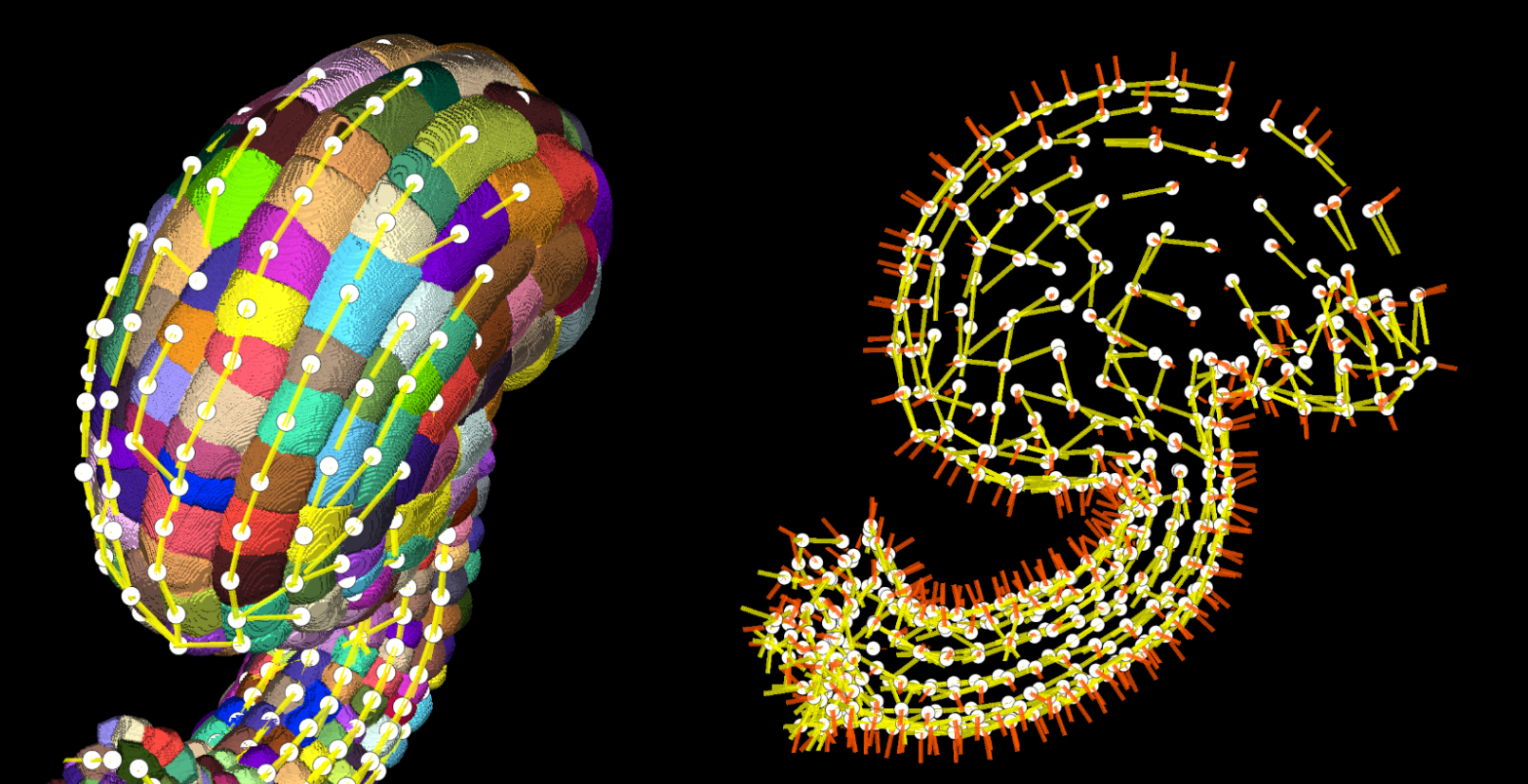}
  \caption{Automatically extracted local orientation features: the growth and surface directions are represented by the solid yellow lines and solid red lines respectively. Left: a 3D surface view of a mature Arabidopsis ovule overlaid with the predicted growth directions. The axis predictions align well with the filar arrangement of cells on the organ surface. Right: surface and growth orientations from the same ovule. Although the orientation of the growth axis follows the regular structure of the organ, the direction of the axis is arbitrary.}
  \label{fig:per_anti}
\end{figure}

Moreover, these tissues (\textit{L1} to \textit{L4}), follow a stratified pattern as they continue dividing anticlinally. In this case, we propose another simple heuristic to construct a \textit{surface axis}, parallel to the stratification direction.
Both algorithms are described in details in \cref{suppl:sec2}.
The approximate \textit{growth axis} and \textit{surface axis} we find in this way have two significant limitation: the axes are not guaranteed to be orthogonal, and only their orientation is defined, but not their direction. Lastly, in order to have a complete 3D basis for our local reference system, we simply compute a third axis orthogonal to the first two. An illustration of the predicted \textit{growth} and \textit{surface axes} can be found in \cref{fig:per_anti}.

\subsection{Feature extraction}
\label{subsec:basic-feat}
As a first step, we compute the Cell Adjacency Graph $G\left( \mathcal{N}, \mathcal{E} \right)$, where every cell is represented as node $n_i \in \mathcal{N}$ and an edge $e_{i,j} \in \mathcal{E}$ connects every pair of touching cells.
Secondly, we sample a fixed number of points from the surface of each cell. To obtain equally distributed points, we use the farthest points sampling algorithm \cite{qi2017pointnet++} over the cell surfaces.
These sampled points are used to compute downstream features more efficiently, when necessary. However, the same surface samples could potentially be used to learn features end-to-end, using ideas from equivariant point cloud architectures. Although this direction is enticing, it would result in a much more complex pipeline.
We also sample points from the boundary shared by any two touching cells, using the same sampling strategy. In each case, we sample 500 points per node/per edge.

We can divide all features into two categories: \textit{invariant} and \textit{covariant features}.
\paragraph{Invariant features} are independent from the reference system, and do not change under rotation or translation of the specimen. These features are either morphological --- such as the cell volume, surface area, lengths along the local reference axis, and the angle formed between the surface and growth axis --- or they are derived from the Cell Adjacency Graph, such as the shortest path from each node to the ovule surface, current-flow closeness centrality \cite{stephenson1989rethinking}, and degree centrality.
\paragraph{Covariant features} instead transform with the reference system under rotation and translation. In this category we include: the cell center of mass, the local reference axes, and the cell PCA axes. In addition we also include: the angles formed between the local and the global reference system, and the angles between PCA axes and the global reference system. Although angles are not formally covariant, they are derived from covariant features and transform predictably under rotation and translation.
\paragraph{Edge features} are also pre-computed in our benchmark, in particular: shared cell boundary surface, the distance between adjacent centers of mass, and the angles between the local reference axes of adjacent cells. But as discussed in \cref{subsec:res}, these brought no significant improvement in the architectures we tested.

\subsection{Feature homogenization} 
Appropriate feature processing has a great impact on the training dynamics. It is extremely important to properly normalize and homogenize incommensurable features before concatenating them.

To provide the orientation of PCA and local reference system axes to the network, we used the following transformation: 
\begin{equation*}
f: \mathbb{R}^3 \rightarrow \mathbb{R}^6, \left( x, y, z \right) \rightarrow \left( x^2, y^2, z^2, xy, xz, yz \right),    
\end{equation*}
That is, we removed the ambiguity in choosing direction and not just orientation. This yields an embedding of the manifold of orientations ${\mathbb{R}P}^2$ in $\mathbb{R}^6$, whose existence is guaranteed by the Whitney embedding theorem \cite{whitney1944self, adachi2012embeddings}.

All scalar features (except angles) are normalized to zero mean and unit variance. Furthermore, we encode categorical features as a one-hot vector and scale vector features to unit norm.

This protocol has been established by testing different preprocessing and normalization strategies for each group of features. Results are shown in \cref{suppl:sec3}.

We made a best effort to ensure that the features we selected are expressive. However, our data loader can be easily extended to add new features or change the processing of the current features. The features discussed so far are all included in the loader by default, but in \cref{suppl:sec3} we list additional features. All raw segmentations, annotations and pre-computed features are available at \url{https://zenodo.org/record/6374104}.

\section{Benchmarking graph neural networks}
\label{sec:baseline}
To show the accuracy of existing methods on this new benchmark, we experiment with a large variety of models. We tried to represent the most prevalent graph neural networks paradigms. In particular, we tested graph convolutional networks, attention or transformer based architectures as well as message-passing based architectures.

\paragraph{Architectures:} We tested the following graph neural network architectures: \textit{GCN} \cite{kipf2016semi}, \textit{GraphSAGE}, \cite{hamilton2017inductive}, \textit{GIN} \cite{xu2018powerful}, \textit{GCNII} \cite{chen2020simple}, and \textit{DeeperGCN} \cite{li2019deepgcns, li2020deepergcn}; all as implemented in \cite{Fey2019}. Moreover, we tested a two layer re-implementation of the graph attention network \textit{GAT} \cite{velickovic2018graph} architecture, a variation of the same architecture \textit{GATv2} using the graph convolution operation introduced in \cite{brody2021attentive}, and a further variation \textit{TransformerGCN} utilizing a transformer \cite{vaswani2017attention} based graph convolution from \cite{shi2020masked}. 

Two of the architectures tested can take into account edge features, the \textit{TransformerGCN} and the \textit{DeeperGCN}. When trained with edge features we will refer to them as \textit{EdgeTransformerGCN} and \textit{EdgeDeeperGCN}.
All models used softmax as last layer activation. 

\paragraph{Training:} All models have been trained with the ADAM optimizer \cite{kingma2015adam} and cross entropy loss with $\mathbbm{L}_2$ weight penalty. Learning rate, weight regularization and model specific hyperparameters such as: number of layers, hidden features, dropout etc. have been tuned for each model using a coarse grid search.

Every training instance uses a single GPU and less than 2GB of VRAM. A complete five fold cross validation of the \textit{DeeperGCN} completes in under one hour using a single Nvidia RTX6000 GPU. For all experiments, source code is provided at \url{https://github.com/hci-unihd/plant-celltype}.

\begin{table}
  \centering
  \begin{tabular}{@{}lcc@{}}
    \toprule
    Model \\
    \midrule
    Node Classification & top-1 acc. & class-avg. acc.\\
    \midrule
    GIN \cite{xu2018powerful} & $0.714 \pm 0.071$ & $0.563 \pm 0.136$ \\
    GCN \cite{kipf2016semi} & $0.762 \pm 0.043$ & $0.617 \pm 0.077$ \\ 
    GAT \cite{velickovic2018graph} & $0.824 \pm 0.033$ & $0.705 \pm 0.084$ \\
    GATv2 \cite{brody2021attentive} & $0.855 \pm 0.041$ & $0.757 \pm 0.087$ \\
    GraphSAGE \cite{hamilton2017inductive} & $0.859 \pm 0.048$ & $0.765 \pm 0.093$ \\
    GCNII \cite{chen2020simple} & $0.863 \pm 0.050$ & $0.772 \pm 0.100$ \\
    Transf. GCN \cite{shi2020masked} & $0.868 \pm 0.045$ & $0.779 \pm 0.098$ \\
    EdgeTransf. GCN \cite{shi2020masked} &$0.868 \pm 0.044$ & $0.777 \pm 0.098$ \\
    DeeperGCN \cite{li2020deepergcn} &  $\textbf{0.877} \pm \textbf{0.050}$ & $\textbf{0.796} \pm \textbf{0.098}$ \\ 
    EdgeDeeperGCN \cite{li2020deepergcn} & $\textbf{0.878} \pm \textbf{0.047}$ & $\textbf{0.797} \pm \textbf{0.095}$ \\  
    \midrule
    Expert Biologist & $0.932 \pm 0.025$ & $0.909 \pm 0.049$ \\ 
    \bottomrule
  \end{tabular}
  \caption{Top-1 accuracy and class-average accuracy obtained by different architecture on the node classification task. \textit{DeeperGCN} is consistently the best performing architecture. Nevertheless, \textit{DeeperGCN} results are still weaker than human expert. \textit{EdgeDeeperGCN} and \textit{EdgeTransformerGCN} have been trained with additional edge features, but their impact on our metrics was not significant. Uncertainty is defined as the standard deviation over all specimens and cross-validation folds.}
  \label{tab:node_class_eval}
\end{table}

\subsection{Baseline results}
\label{subsec:res}
In \cref{tab:node_class_eval}, we present the best performing models over the five-fold cross-validation splits according to the class-average accuracy. \textit{DeeperGCN} is consistently the best performing architecture, although it does not match yet the expert biologist performance.

\begin{figure}
    \centering
    \includegraphics[width=1\linewidth]{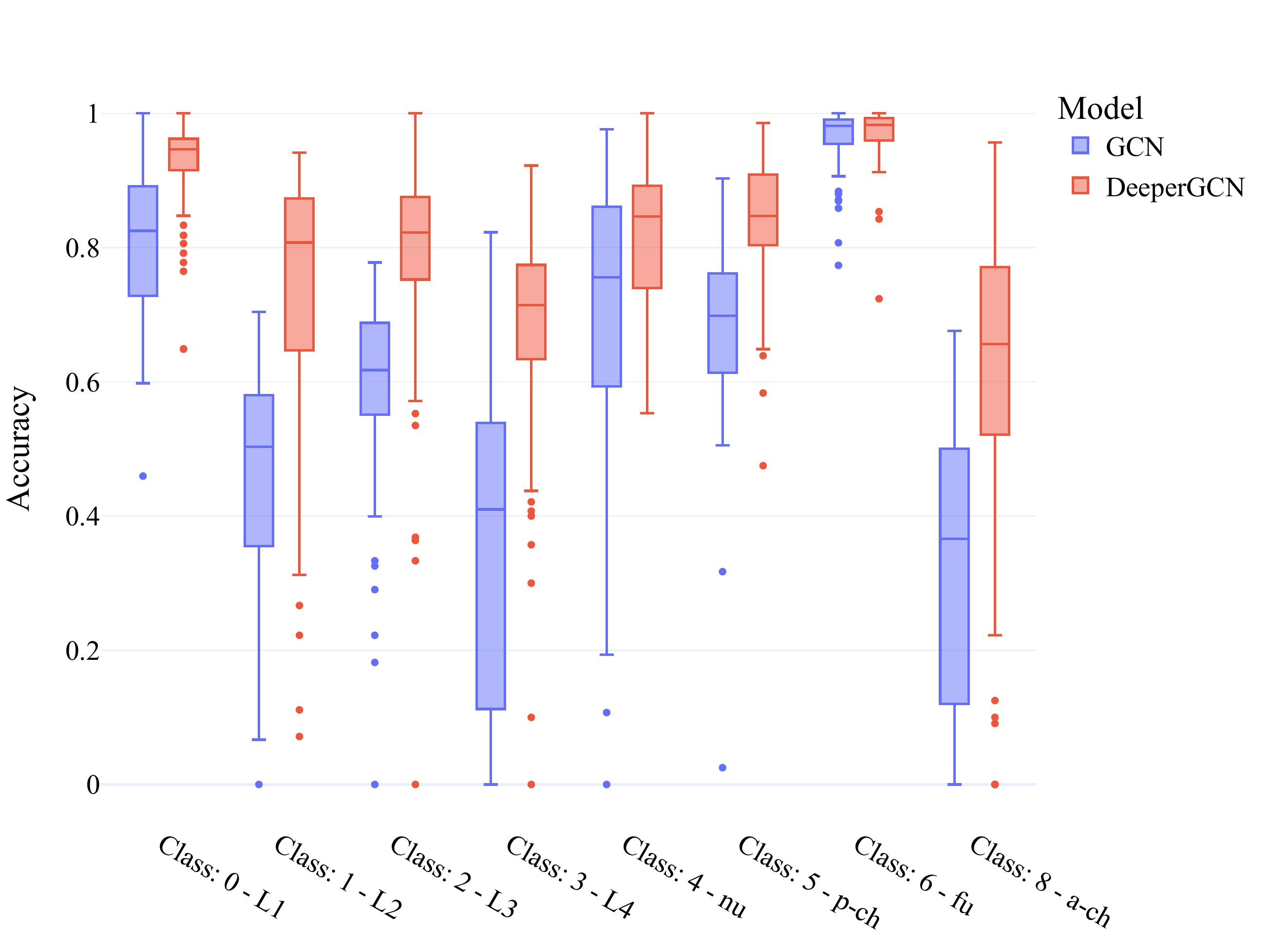}
  \caption{Top-1 accuracy divided by ground truth category. Accuracy varies drastically across classes. In particular, the \textit{GCN} accuracy plummets for \textit{L2}, \textit{L4}, and \textit{a-ch} tissues.}
  \label{fig:classes_accuracy}
\end{figure}

\begin{figure*}[t!]
  \centering
   \includegraphics[width=1\linewidth]{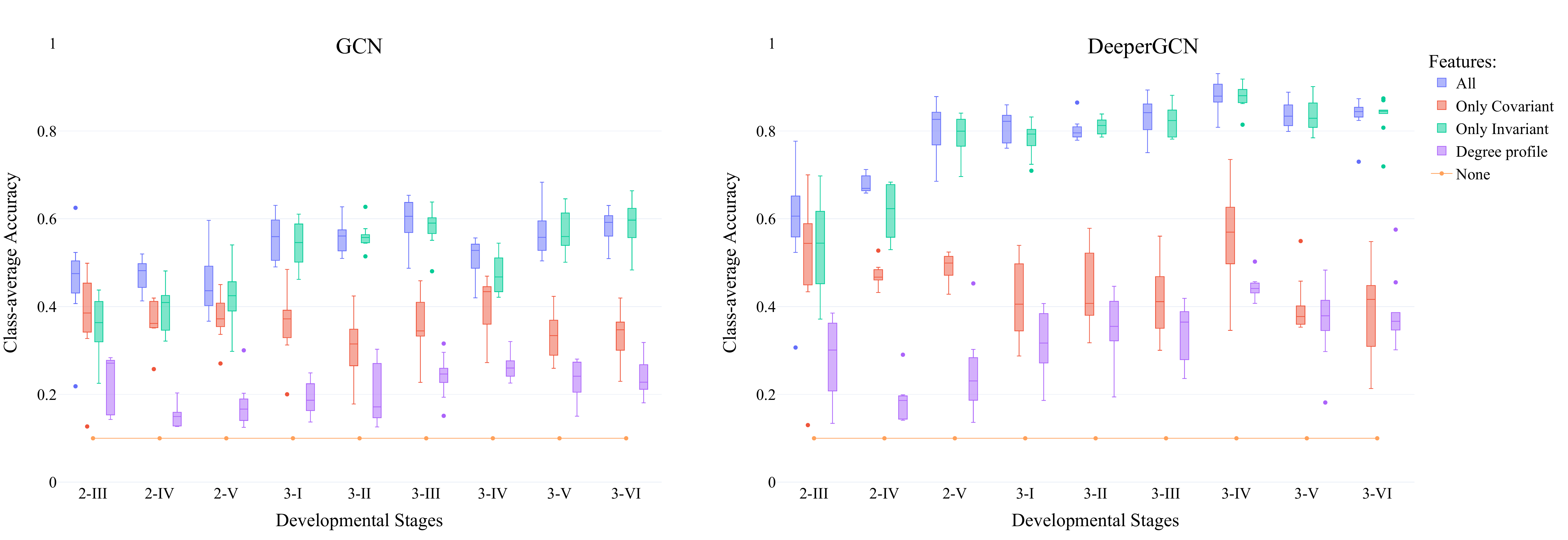}
    \caption{Comparison of the class average accuracy for four different feature sets. As one would expect, using all features consistently achieves the highest accuracy. Interestingly, one can observe that the covariant features mostly contribute to accuracy in the early stages (2-III to 2-V). In the later stages (3-I to 3-VI), the invariant features are the major contributors to the overall accuracy. Lastly, even when using only the local degree profile of a node as its feature, the networks are still able to make better than random predictions on the task.}
  \label{fig:feat_imp}
\end{figure*}

Overall, all attention mechanism methods performed relatively well, while simpler models like \textit{GCN} and \textit{GIN} struggled the most.
For both high- and low-end models, using edge features did not increase accuracy.
\cref{fig:feat_imp} shows the class-average accuracy broken down by developmental stages. While \textit{DeeperGCN} has a decisive advantage in the later stages compared to \textit{GCN}, we can observe a much narrower spread in the early stages (2-III to 2-V).
\cref{fig:classes_accuracy} shows the average accuracy by cell type. One can easily identify the categories \textit{L2}, \textit{L4} and \textit{ac} to be the most challenging for both \textit{GCN} and \textit{DeeperGCN}.

\subsection{Additional experiments}
\label{additiona_exp}
The same standardized setup was used for all our additional experiments. We limited ourselves to two models, the \textit{GCN} \cite{kipf2016semi} and the \textit{DeeperGCN} \cite{li2020deepergcn}, and used optimal hyper-parameters from \cref{sec:baseline}, see \cref{suppl:sec4} for each.

\subsubsection{Importance of the global reference system}
\label{subsec:grs_importance}
As discussed in \cref{subsec:grs}, some of the features are sensitive to the orientation of the specimen. We tested and compared four landmark-based orientations plus the trivial orientation, i.e., center of mass in the origin and original orientation as acquired by the microscope. From \cref{fig:grs_stability} we can observe no significant difference between different orientations.
The only relevant difference is that the reference system found using the labels exhibits an overall smaller variance across specimens. On the other end, the trivial representation manifests the most significant number of outliers. 
Another interesting question for practitioners is the ability to generalize if the reference system changes. 
We tested that by training our pipeline with three different orientations and always testing on the same. 
As shown in \cref{tab:generalization}, at test time, there is a significant drop in accuracy when evaluating on a different orientation. However, the experiment also shows that this can be mitigated by training on multiple orientations at the same time. 

\subsubsection{Invariant vs covariant features}
To evaluate the respective contribution of invariant and covariant features to the accuracy, we trained our models using solely one of the two. \cref{tab:features_importance} shows that the invariant features are unquestionably what the neural networks rely on most.

To further understand the difference, in \cref{fig:feat_imp} we report our results for each development stage separately. One can see a distinction between the two feature groups: the invariant features have a stronger impact on the accuracy in the later stages (3-I to 3-VI), while the covariant features have a more prominent role in the early stages (2-III to 2-V).

\subsubsection{Local graph features alone}
To emphasize the importance of the Cell Adjacency Graph structure for solving this task, we also trained models using only the degree of a node and distribution of degrees of is neighbours, the so called degree profile \cite{cai2018simple}.
This is an extremely unfavorable setup since the degrees are quite homogeneous across the ovule cell graph. The results, shown in \cref{fig:feat_imp}, reveal that even then, the networks are still able to predict systematically better than chance. Moreover, \textit{DeeperGCN} in later stages performed comparably to the same model trained using the covariant features.

\begin{table}
  \centering
  \begin{tabular}{@{}lcc@{}}
    \toprule
    Features & GCN                      & DeeperGCN \\
             & $\Delta$ class-avg. acc. & $\Delta$ class-avg. acc. \\
    \midrule
    Only invariant     & $ -0.028 $ & $ -0.019 $  \\
    Only covariant     & $ -0.175 $ & $ -0.337 $  \\
    \bottomrule
  \end{tabular}
  \caption{Contribution of each feature group to the accuracy. The table shows the difference between the class average accuracy of the baseline model and the class average accuracy of the perturbed model. The lower the delta, the higher is the feature importance for the neural networks. For both architectures, the invariant features are substantially more important than the covariant features.}
  \label{tab:features_importance}
\end{table}

\begin{table}
  \centering
  \begin{tabular}{@{}lcc@{}}
    \toprule
    Training Data & GCN & DeeperGCN \\
                  & class-avg. acc.      & class-avg. acc.  \\
    \midrule
    \textit{ES PCA}             & $ 0.578 \pm  0.089$ & $ 0.754 \pm  0.118$ \\
    All but \textit{Label Surf} & $ 0.618 \pm  0.086$ & $ 0.787 \pm  0.103$ \\
    \textit{Label Surf}         & $ 0.613 \pm  0.080$ & $ 0.790 \pm  0.100$ \\
    \bottomrule
  \end{tabular}
  \caption{Generalization under different orientations. We tested how robust our baseline is to changes in the global orientation axes. We trained our model in different orientations, and we tested it on a particular one, \textit{Label Surf}. From the class average accuracy, we can observe a substantial accuracy drop when training on a single different reference system, \textit{ES PCA}. However, this can be easily compensated by augmenting the training data with multiple orientations. Uncertainty is defined as the standard deviation over all specimens.}
  \label{tab:generalization}
\end{table}

\section{Limitations}
\label{sec:limitations}
The CellTypeGraph Benchmark is limited by the number of specimen. Imaging, segmenting, and annotating large 3D volumes is very time-consuming, and thus the number of specimens available is smaller compared to other datasets \cite{wu2018moleculenet, gomez2018automatic}. This limitation can be mitigated by data augmentation. Although not exhaustive, this route did not show visible improvements in our experiments. A description of the experiments performed and results can be found in \cref{suppl:data_aug}.

Nevertheless, the structural complexity of the ovules makes them a perfect candidate for a benchmark. However, ovules are not a universal proxy for all organs in plant-biology, thus pre-trained models on our benchmark are likely to fail on different organs. Possible solutions are: i) retraining the models when labels are available. ii) Using self-supervised methods such as graph autoencoders \cite{kipf2016variational, salehi2020graph} followed by community clustering in latent space. Such a setup has proven successful in single-cell data analysis \cite{amodio2019exploring, lopez2017deep}. iii) Speeding up semi-automated labeling by employing active learning.

In the benchmark, the two main sources of complexity are the hand-crafted features and the need to define a global orientation. End-to-end feature learning could be achieved by applying models from the point clouds geometric deep learning domain \cite{qi2017pointnet, qi2017pointnet++, li2018pointcnn} to our surface sample, see \cref{subsec:basic-feat}. Moreover, new exciting frameworks like \cite{e3nn} are expediting easy access to equivariant neural network architectures.

\section{Acknowledgments}
%All authors are deeply grateful to Tejasvinee Mody for the immense labeling and data proofreading efforts.
This work was supported by the Deutsche Forschungsgemeinschaft (DFG) research unit FOR2581 Quantitative Plant
Morphodynamics.

\section{Conclusion}
We have introduced a new graph benchmark for node classification, extensively tested several relevant graph neural networks architectures, and released tools for quick experimentation, data handling, and evaluation.

Although results from our baseline are encouraging, we are excited to discover how far the community will push this task. As discussed in \cref{sec:limitations}, tissue labeling in 3D virtual organs is a relevant and relatively unexplored domain. We also hope to see use of the benchmark outside the strict paradigms of supervised learning, such as self-supervised and active learning.
%%%%%%%%% REFERENCES
{\small
\bibliographystyle{ieee_fullname}
\bibliography{bibliography_celltypegraph}
}

\clearpage
\appendix
\section{Global reference axis}
\label{suppl:sec1}
% Heuristics Description
Our landmark-based global reference systems are defined by fixing an origin and an orthogonal 3D vectors basis. A good land-mark should be retrievable from the ground truth labels at training time, but should also be easy to identify by a human without ground truth labels at test time.

We choose the following four approaches: 

\paragraph{Label Surf:} The origin is defined as the center of mass of the \textit{L1} tissue. 
The main axis is found by approximately determining the integument tissue symmetry axis. This is achieved by finding for each tissue in the integument (\textit{L2, L3, L4, es, nu}) the respective center of mass. Then, we use least-square linear regression to find the best line interpolating the integument tissue. The second axis is found by finding the line passing through \textit{fu} tissue center of mass and perpendicular to the main axis. The third axis is computed by taking the cross-product between the first and second axes.

\paragraph{Label Fu:} The origin is defined as the center of mass of the \textit{fu} tissue. The global axes are computed as in the \textit{Label Surf}.

\paragraph{Es trivial:} Here we use the \textit{Es} tissue to fix an origin for the reference system. It is usually easily identifiable by its large size and central position. We set the reference system origin to the  \textit{Es} center of mass. While for the axis, we use the original orientation as acquired by the microscope. 

\paragraph{Es PCA:} The origin is the same as \textit{Es Trivial}. But the system axes are set to the PCA axes of the whole ovule.\\

Our python implementation can be found at: \url{https://github.com/hci-unihd/plant-celltype/blob/main/plantcelltype/features/cell_vector_features.py}.\\

\section{Growth and surface axis}
\label{suppl:sec2}
To compute the local reference system, we designed two simple heuristics. 
We estimate the surface axis of a cell by averaging over directions corresponding to the edges connecting the cell to all neighbors which are closer to the surface. These directions are defined as the edge surface normal direction.\\ 
The growth axes are found by looking for each cell the most co-linear pair of neighboring cells.

The algorithms used to compute the surface axis and growth axis are described in more detail respectively in \cref{alg:surf} and \cref{alg:growth}. Our python implementation can be found at:
\url{https://github.com/hci-unihd/plant-celltype/blob/main/plantcelltype/utils/axis_transforms.py}.

\begin{algorithm*}
\begin{algorithmic}
\Require node, edges \Comment{Vectors containing: nodes ids, edges ids.}
\Require hops \Comment{Vectors containing: number of hops from each node to the surface.}
\Require directions, bg \Comment{Vector containing: edges directions (surface normal), background node id.}
\State N $\gets$ \textit{len}(node)
\State surface-axis $\gets$ \textit{zeros}(N, 3) \Comment{Initialize an array full of zeros.}

\For{($i=0$, $i=\text{N}$, $i$\texttt{++})}
\If{$\text{hops}_i = 1$} \Comment{I.e. node is on the organ surface.}
    \State e $\gets$ \textit{find-edge}($\text{node}_i$, bg, edges) \Comment{Find edge id between $\text{node}_i$/$\text{node}_j$.}
    \State $\text{surface-axis}_i \gets \text{directions}_{e}$
\Else
    \State $\text{neighbors}_i \gets$ \textit{find-neighbors}($\text{node}_i$, edges) \Comment{Find all nodes neighbors of $\text{node}_i$.}
    \State $\text{N}_i, \text{count} \gets$ $\textit{len}(\text{neighbors}_i), 0$
    \For{$j=0$, $j=\text{N}_i$, $j$\texttt{++}} \Comment{Loop over the neighborhood.}
    \If{$\text{hops}_j < \text{hops}_i$} \Comment{If neighbor is closer to surface.}
        \State e $\gets$ \textit{find-edge}($\text{node}_i$, $\text{node}_j$, edges) \Comment{Find edge id between $\text{node}_i$/$\text{node}_j$.}
        \State $\text{surface-axis}_i \gets \text{surface-axis}_i + \text{directions}_{e}$ \Comment{Average edges normal.}
        \State count = count + 1
    \EndIf
    \State $\text{surface-axis}_i \gets \text{surface-axis}_i / \text{count}$
    \EndFor
\EndIf
\EndFor
\end{algorithmic}
\caption{Compute Surface Axis}\label{alg:surf}
\end{algorithm*}

\begin{algorithm*}
\begin{algorithmic}
\Require node, edges \Comment{Vectors containing: nodes ids, edges ids.}
\Require coms, hops \Comment{Vectors containing: nodes center of mass, number of hops from each node to the surface.}

\State N $\gets$ \textit{len}(node)
\State  growth-axis $\gets$ \textit{zeros}(N, 3) \Comment{Initialize an array full of zeros.}

\For{($i=0$, $i=\text{N}$, $i$\texttt{++})}
    \State $\Theta_{min} \gets$ 1
    \State $\text{neighbors}_i \gets$ \textit{find-neighbors}($\text{node}_i$, edges) \Comment{Find all nodes neighbors of $\text{node}_i$.}
    \State $\text{N}_i \gets$ $\textit{len}(\text{neighbors}_i)$

    \For{$j=0$, $j=\text{N}_i$, $j$\texttt{++}} \Comment{Loop over all tuple ($\text{node}_j$, $\text{node}_j$) of distinct neighbors of $\text{node}_i$.}
        \State $\text{vector}_{ij}$ = \textit{get-vector}($\text{node}_i$, $\text{node}_j$, coms) \Comment{Find vector connecting $\text{node}_i$/$\text{node}_j$ center of masses.}
        \For{$k=j+1$, $k=\text{N}_i$, $k$\texttt{++}}
            \State $\text{vector}_{ik}$ = \textit{get-vector}($\text{node}_i$, $\text{node}_k$, coms)
            \State $\Theta$ = \textit{get-angle}($\text{vector}_{ij}$, $\text{vector}_{ik}$) \Comment{Angle is measured as the normalized dot product between the two vectors.}
            
            \If{$\Theta < \Theta_{min}$ \textbf{and} $\text{hops}_i = \text{hops}_j = \text{hops}_k$} \Comment{When the angle is minimum the cell are the most co-linear.}
                \State $\Theta_{min} \gets \Theta$
                \State $\text{growth-axis}_i \gets \text{vector}_{ij}$
            \EndIf
        \EndFor
    \EndFor

\EndFor
\end{algorithmic}
\caption{Compute Growth Axis}\label{alg:growth}
\end{algorithm*}

\section{Features}
\label{suppl:sec3}
\begin{figure}
  \centering
  \includegraphics[width=1\linewidth]{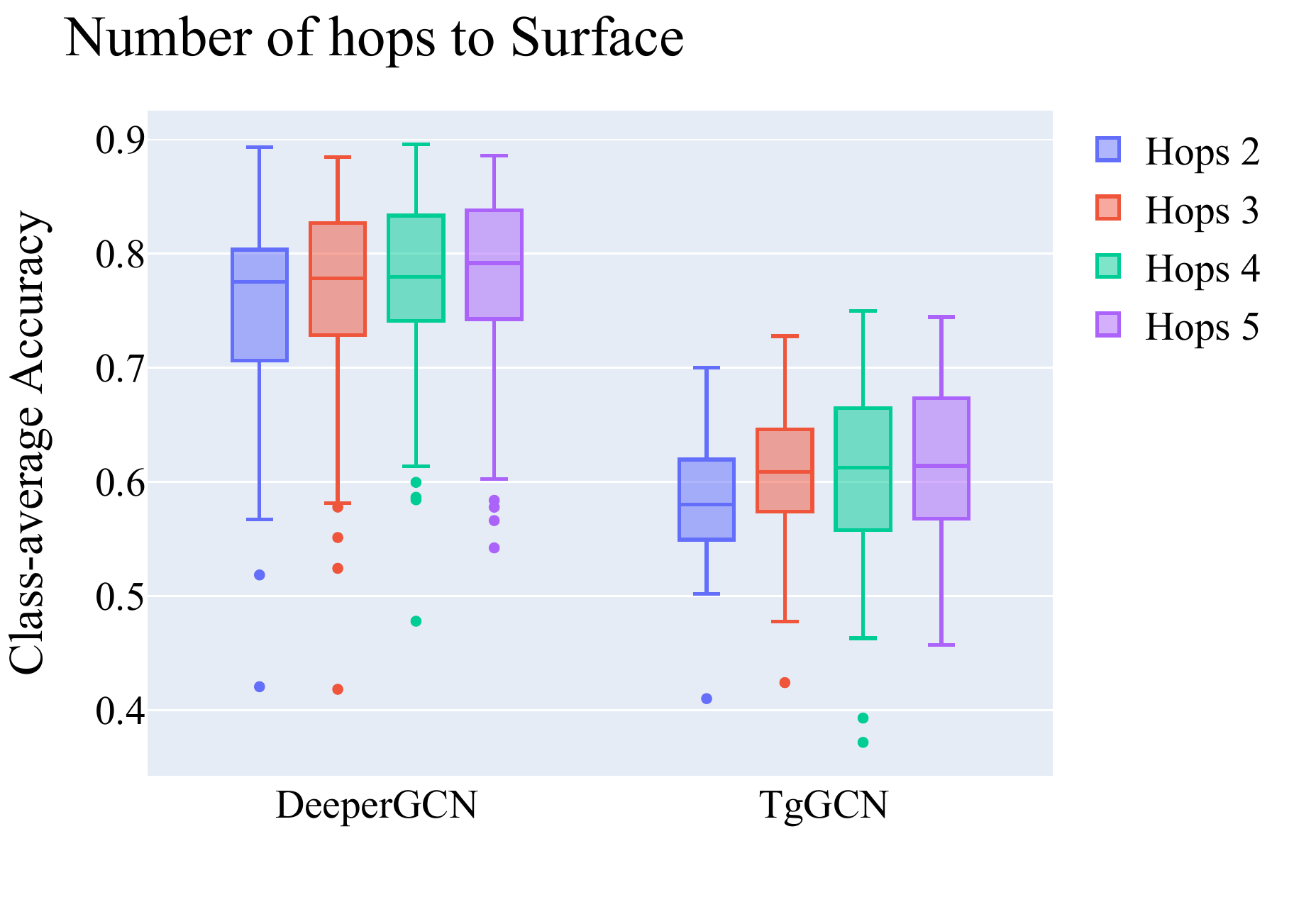}
  \caption{Class-average accuracy with varying maximum number of hops. Hops to surface are intuitively closely related to the tissue stratification \textit{L1-L4}, but their relation loosen with depth. We tested how important this feature is by clipping its value. From the results one can see that the feature contribution saturates after three hops.}
  \label{fig:hops}
\end{figure}

\begin{figure}
  \centering
  \includegraphics[width=1\linewidth]{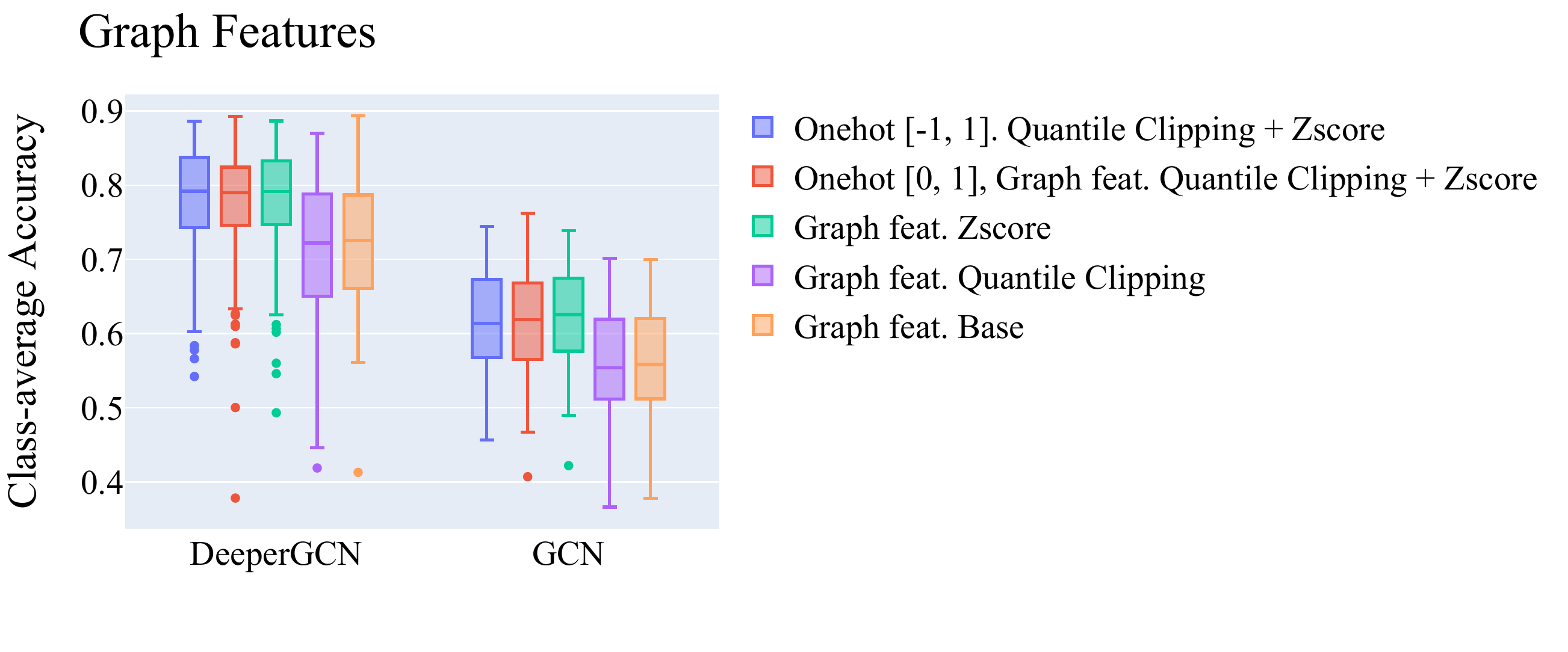}
  \caption{Class-average accuracy between different graph feature normalizations. One can see a slight accuracy improvement using z-score normalization.}
  \label{fig:graph}
\end{figure}

\begin{figure}
  \centering
  \includegraphics[width=1\linewidth]{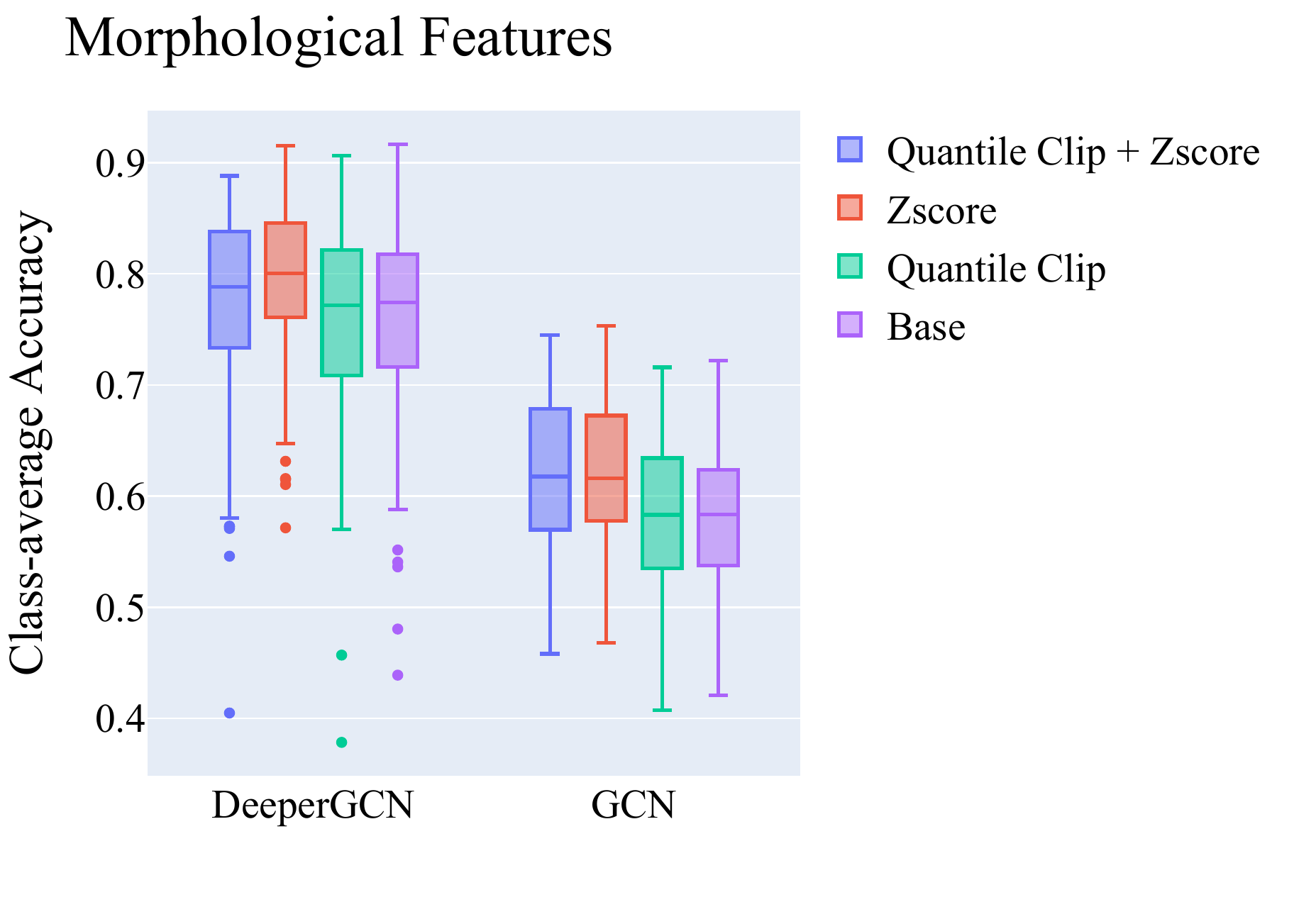}
  \caption{Class-average accuracy comparison between different morphological features normalizations. One can see a slight accuracy improvement using z-score normalization.}
  \label{fig:morph}
\end{figure}

\begin{figure}
  \centering
  \includegraphics[width=1\linewidth]{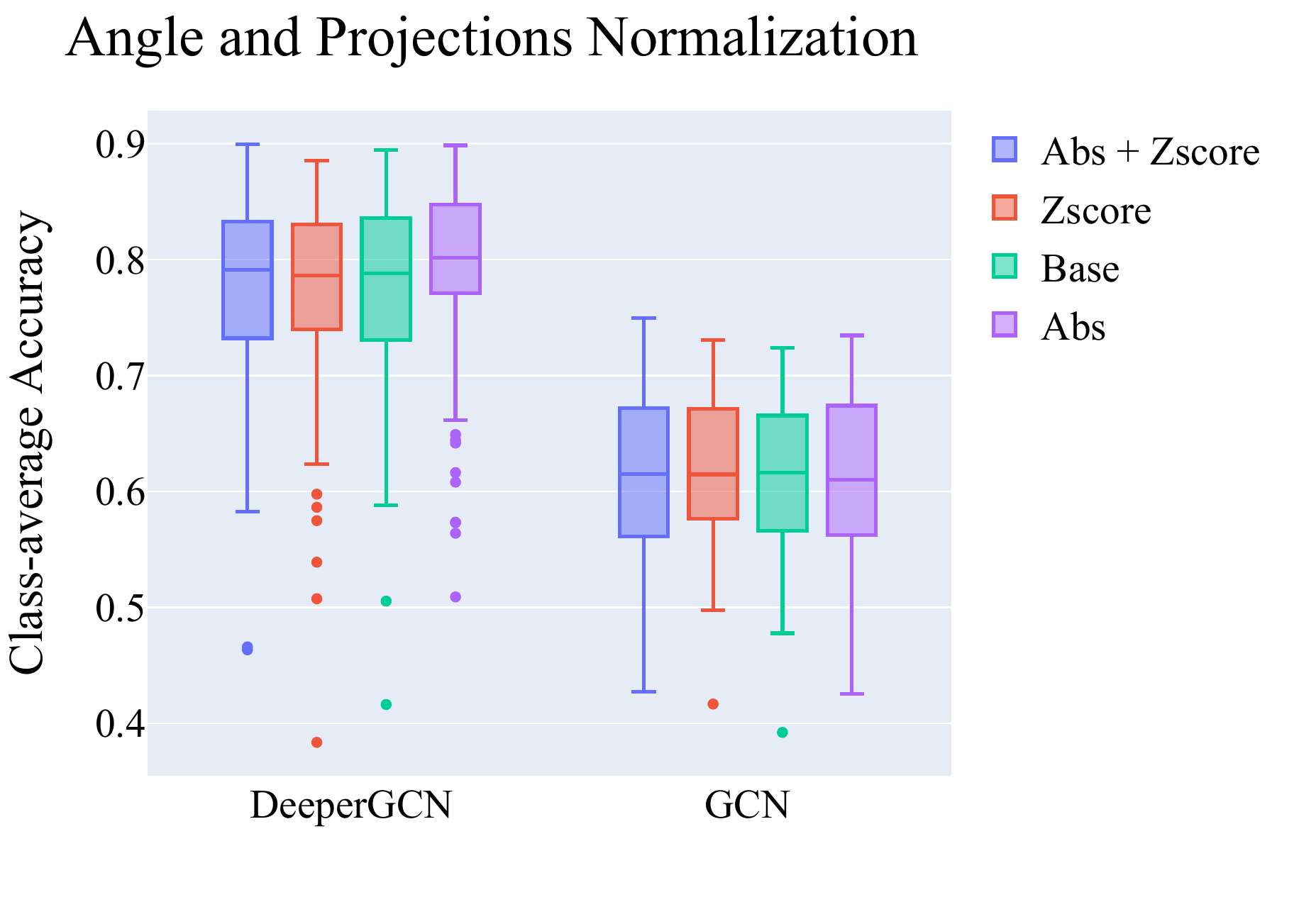}
  \caption{Class-average accuracy between different angles normalization. Angles are naturally normalized between -1 and 1, in our experiments the z-score had no significant impact.}
  \label{fig:angle}
\end{figure}

\begin{figure}
  \centering
  \includegraphics[width=1\linewidth]{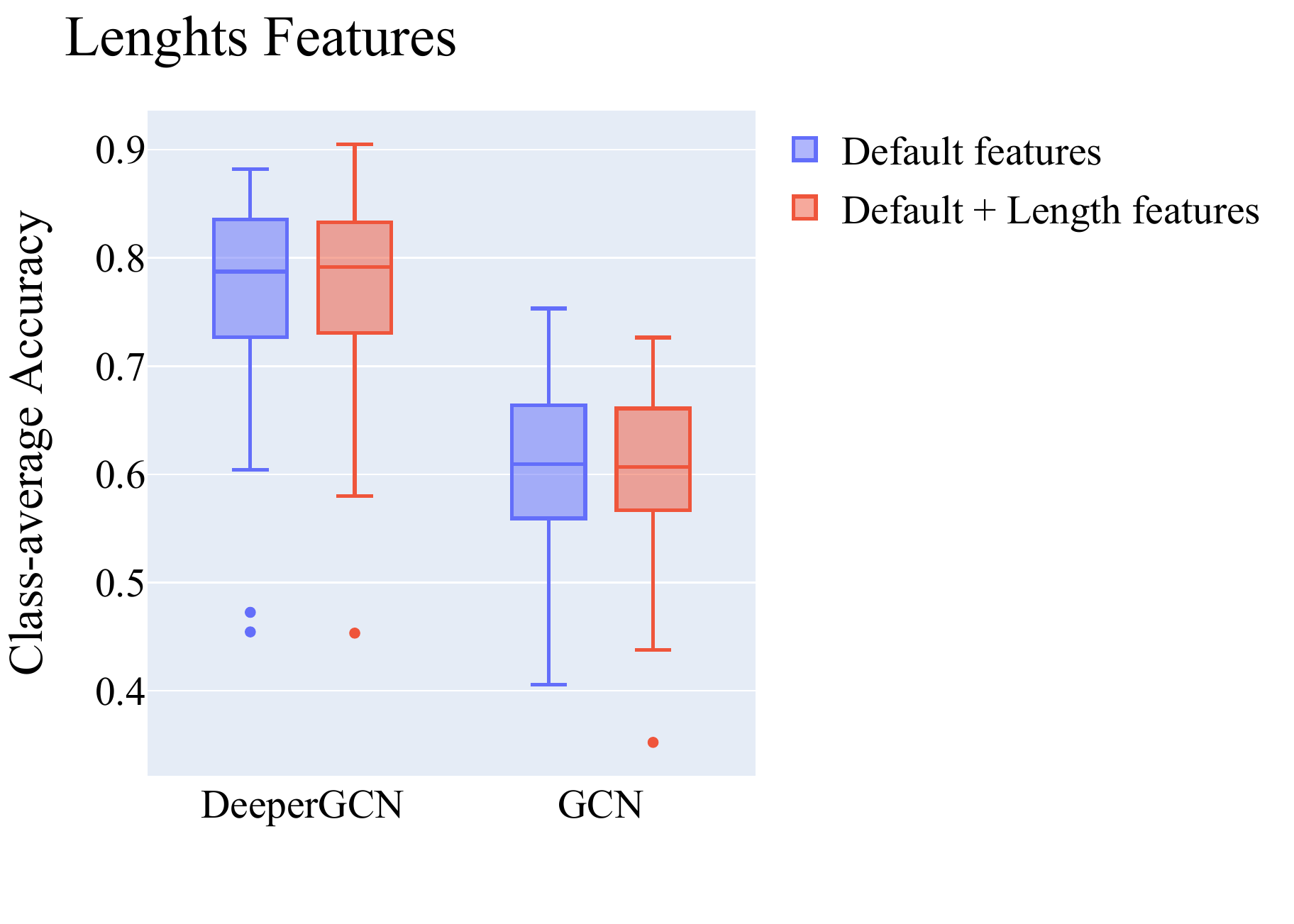}
  \caption{Class-average accuracy between: baseline features only, and baseline features plus additional lengths features. In our experiments the additional lengths showed no significant contribution to the accuracy.}
  \label{fig:lengths}
\end{figure}

We here report the results of additional experiments conducted to identify the best feature homogenization strategy. In \cref{fig:orientation}, we tested the difference between different vector representations. In \cref{fig:graph} and  \cref{fig:hops}, we tested the normalizations for the graph features and the maximum value to be used in our hops to surface feature. In \cref{fig:morph} and \cref{fig:angle} we tested different normalization applied to respectively morphological and angles features. Lastly, in \cref{fig:lengths} we tested the impact of using lengths measured in several directions as features, similarly to \cite{schmidt2014irocs}.
An overview of all features available in the CellTypeGraph benchmark is reported in \cref{tab:full_node_features}, \cref{tab:full_edge_features}. Our python implementation can be found at: \url{https://github.com/hci-unihd/plant-celltype/tree/main/plantcelltype/features}

\begin{figure}
  \centering
  \includegraphics[width=1\linewidth]{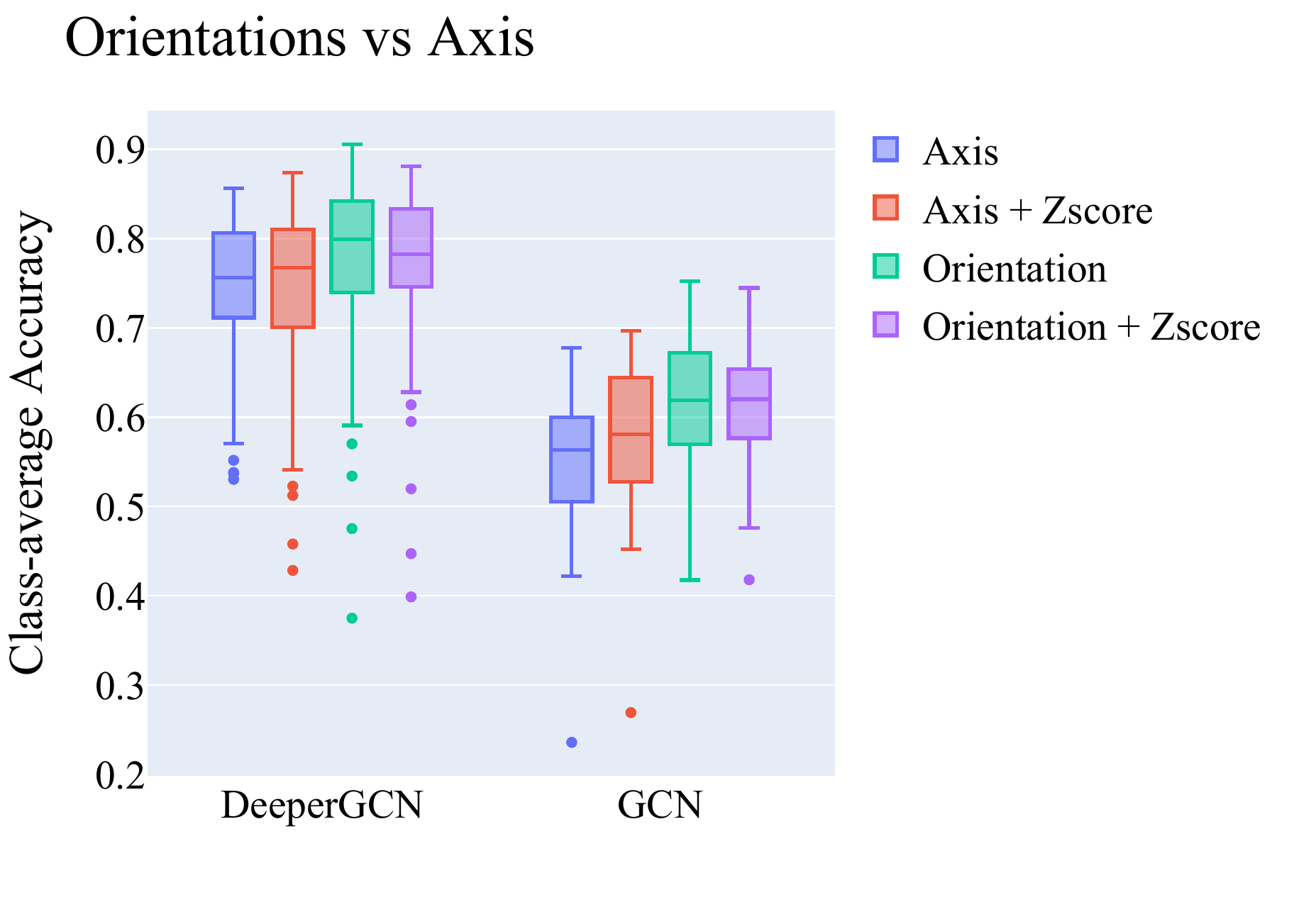}
  \caption{Class-average accuracy comparison between different vector features representations. Using the orientation instead of axis resulted in a small but consistent improvement in performance.}
  \label{fig:orientation}
\end{figure}

\begin{table*}[h]
  \centering
  \begin{tabular}{@{}lcccp{.4\textwidth}}
    \toprule
    Feature Name               & Invariant & Default    & Size & Description\\
    \midrule
    Center of mass             &           & \checkmark & 3  & Cell center of mass represented in the global reference system, expressed in $\mu m$.\\
    Center of mass / GRS Proj. &           &            & 3  & Angles between the cell center of mass and the global reference system.\\
    \midrule
    LRS axis                   &           &            & 9  & Growth axis, surface axis and third perpendicular axis.\\
    LRS orientations           &           & \checkmark & 18 & Direction invariant growth axis, surface axis and third perpendicular axis.\\
    LRS / GRS Proj.            &\checkmark & \checkmark & 9  & Angles between the LRS axis and the global reference system.\\
    Growth/Surface axis angle  &\checkmark & \checkmark & 1  & Angle between growth and surface axis.\\
    Growth axis alignment      &\checkmark & \checkmark & 1  & Angle between periclinial cell walls along predicted growth direction, measure how good is the fit is.\\
    Length LRS                 &\checkmark & \checkmark & 3  & Cell length along the LRS directions.\\
    \midrule
    PCA axis                   &           &            & 9  & Principal component analysis axis.\\
    PCA orientations           &           & \checkmark & 18 & Direction invariant principal component analysis.\\
    PCA / GRS Proj.            &\checkmark & \checkmark & 9  & Angles between the PCA axis and the global reference system.\\
    PCA explained variance     &\checkmark & \checkmark & 3  & PCA axis explained variance.\\
    \midrule
    Surface                    &\checkmark & \checkmark & 1  & Cell surface area in $\mu m$.\\
    Volume                     &\checkmark & \checkmark & 1  & Cell volume in $\mu m$.\\
    Lengths uniform samples    &           &            & 64 & Cell length in uniform directions.\\
    \midrule
    Hops to Surface            &\checkmark & \checkmark & 1  & Shortest path length on the graph between a cell and the surface. For this measure we ignore the geo-localization of nodes, and consider all neighbors one hop distant.\\
    Degree Centrality          &\checkmark & \checkmark & 1  & Degree centrality.\\
    CFC centrality             &\checkmark & \checkmark & 1  & Current-flow closeness centrality.\\
    \bottomrule
  \end{tabular}
  \caption{Complete list of node features pre-computed in the CellTypeGraph.}
  \label{tab:full_node_features}
\end{table*}

\begin{table*}
  \centering
  \begin{tabular}{@{}lcccp{.4\textwidth}}
    \toprule
    Feature Name               & Invariant & Default    & Size & Description\\
    \midrule
    Center of mass surface     &\checkmark & \checkmark & 3  & Cell Surface (edge) center of mass represented in the global reference system, expressed in $\mu m$.\\
    Center of mass distance    &\checkmark & \checkmark & 1  & Distance between two adjacent cell center of mass. \\
    Center of mass /GRS Proj.  &           & \checkmark & 3  & Angles between two adjacent cell center of mass direction and the global reference system.\\
    \midrule
    LRS Proj.                  &\checkmark & \checkmark & 3  & Angles between local reference system in adjacent cells.\\
    \midrule
    Surface                    &\checkmark & \checkmark & 1  & Edge surface area in $\mu m$.\\
    \bottomrule
  \end{tabular}
  \caption{Complete list of edge features pre-computed in the CellTypeGraph.}
  \label{tab:full_edge_features}
\end{table*}

\section{Grid search complete results}
\label{suppl:sec4}
% a big table here with all results from the grid search and normalization
Experiment parameters setup is reported in \cref{tab:full_results}. The configuration files used to run our experiments can be fount at: \url{https://github.com/hci-unihd/plant-celltype/tree/main/experiments/node_grid_search}
\begin{table*}[h]
  \centering
  \begin{tabular}{@{}lllcc@{}}
    \toprule
    Model & Optimizer     & Model Params.  & top-1 acc. & class-avg. acc.\\
    \midrule
    GIN \cite{xu2018powerful}  & lr $=10^{-2}$ & \# feat $= 64$ & $0.714 \pm 0.071$ & $0.563 \pm 0.136$ \\
          & wd $=10^{-5}$ & \# layers $= 2$ &\\
          &               & dropout $= 0.1$   &\\
    \midrule
    GCN  \cite{kipf2016semi} & lr $=10^{-2}$ & \# feat $= 128$ & $0.762 \pm 0.043$ & $0.617 \pm 0.077$ \\
          & wd $=10^{-5}$    & \# layers $= 2$ &\\
          &               & dropout $= 0.5$   &\\
    \midrule
    GAT \cite{velickovic2018graph}  & lr $=10^{-3}$ & \# feat $= 256$ &$0.824 \pm 0.033$ & $0.705 \pm 0.084$ \\
          & wd $=0$ & \# layers $= 2$ &\\
          &               & dropout $= 0.5$   &\\
          &               & heads $= 3$       &\\
          &               & concat. $=$ True  &\\
    \midrule
    GATv2 \cite{brody2021attentive} & lr $=10^{-3}$ & \# feat $= 256$ &$0.855 \pm 0.041$ & $0.757 \pm 0.087$ \\
          & wd $=10^{-5}$       & \# layers $= 2$ &\\
          &               & dropout $= 0.5$   &\\
          &               & heads $= 3$       &\\
          &               & concat. $=$ True  &\\
    \midrule
    GraphSAGE \cite{hamilton2017inductive} & lr $=10^{-3}$ & \# feat $= 128$ &$0.859 \pm 0.048$ & $0.765 \pm 0.093$ \\
              & wd $==10^{-5}$ & \# layers $= 4$ &\\
              &               & dropout $= 0.1$   &\\
    \midrule
    GCNII \cite{hamilton2017inductive} & lr $=10^{-2}$ & \# feat $= 128$      &$0.863 \pm 0.050$ & $0.772 \pm 0.100$ \\
          & wd $=10^{-5}$ & \# layers $= 4$      &\\
          &               & dropout $= 0.0$        &\\
          &               & share weight $=$ False &\\
    \midrule
    Transf. GCN \cite{shi2020masked} & lr $=10^{-3}$ & \# feat $= 128$      &$0.868 \pm 0.045$ & $0.779 \pm 0.098$ \\
                & wd $=10^{-5}$ & \# layers $= 2$      &\\
                &               & dropout $= 0.5$   &\\
                &               & heads $= 3$       &\\
                &               & concat. $=$ True  &\\
    \midrule
    EdgeTransf. GCN \cite{shi2020masked} & lr $=10^{-3}$ & \# feat $= 128$      &$0.868 \pm 0.044$ & $0.777 \pm 0.098$ \\
                    & wd $=0$ & \# layers $= 2$      &\\
                    &               & dropout $= 0.5$   &\\
                    &               & heads $= 5$       &\\
                    &               & concat. $=$ True  &\\
    \midrule
    DeeperGCN \cite{li2020deepergcn} & lr $=10^{-3}$ & \# feat $= 128$      &$\textbf{0.877} \pm \textbf{0.050}$ & $\textbf{0.796} \pm \textbf{0.098}$ \\ 
              & wd $=0$ & \# layers $= 32$      &\\
              &               & dropout $= 0.0$        &\\
    \midrule
    EdgesDeeperGCN \cite{li2020deepergcn} & lr $=10^{-3}$ & \# feat $= 128$      & $\textbf{0.878} \pm \textbf{0.047}$ & $\textbf{0.797} \pm \textbf{0.095}$ \\
                   & wd $=10^{-5}$       & \# layers $= 16$      &\\
                   &               & dropout $= 0.0$        &\\
    \bottomrule
  \end{tabular}
  \caption{Best performing optimizer and model parameters according to the class-avg. accuracy.}
  \label{tab:full_results}
\end{table*}

\section{Expert agreement}
\label{suppl:experts}
\begin{figure}
  \centering
  \includegraphics[width=1\linewidth]{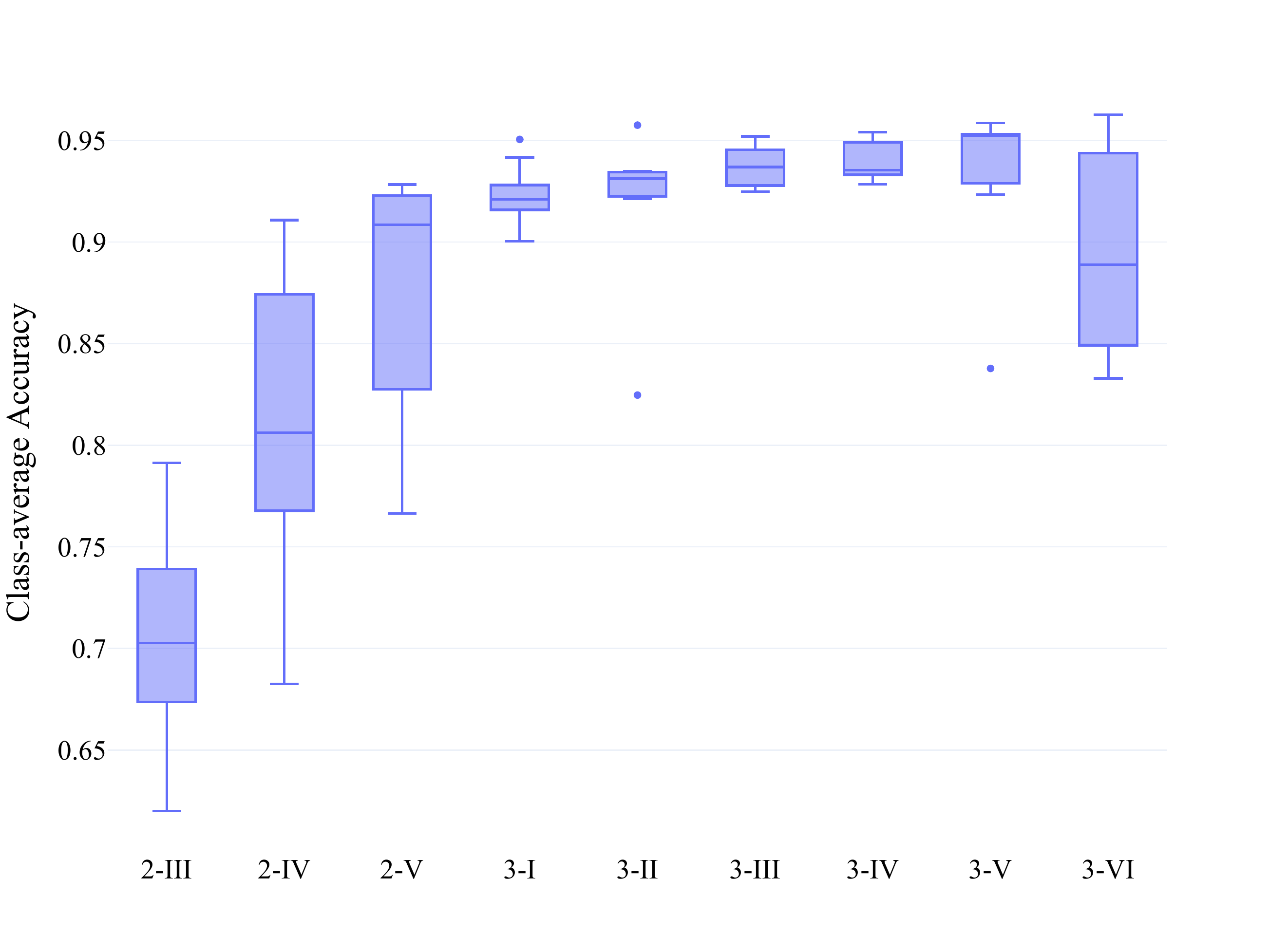}
  \caption{Class-average accuracy obtained by an expert biologist. The early stages pose the most substantial challenges, although the cause of such high variance can be attributed to the smaller number of cells for each specimen.}
  \label{fig:hp_stage}
\end{figure}

\begin{figure}
  \centering
  \includegraphics[width=1\linewidth]{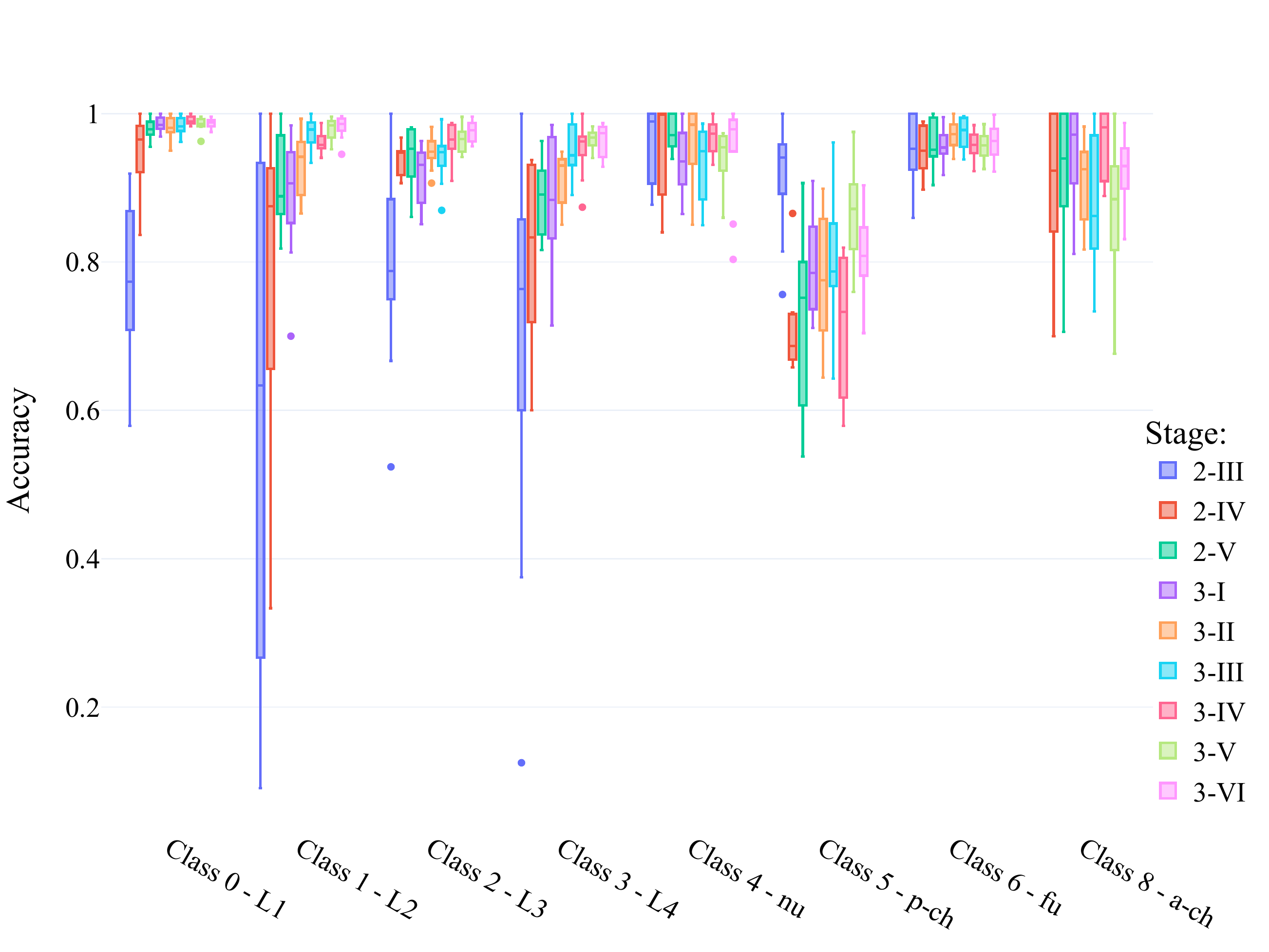}
  \caption{Per-class accuracy obtained by an expert biologist. In late stages the highest variability sources are the cell types \textit{p-ch} and \textit{p-ch}, while for early stages the highest variability is posed by the cell type \textit{L1} to \textit{L4}.}
  \label{fig:hp_class}
\end{figure}

In order to highlights the most challenging aspect of our CellTypeGraph Benchmark, we here report further analysis of the expert biologist performance, see \cref{fig:hp_stage} and \cref{fig:hp_class}.

\section{Data augmentation}
\label{suppl:data_aug}
\begin{figure}
  \centering
  \includegraphics[width=1\linewidth]{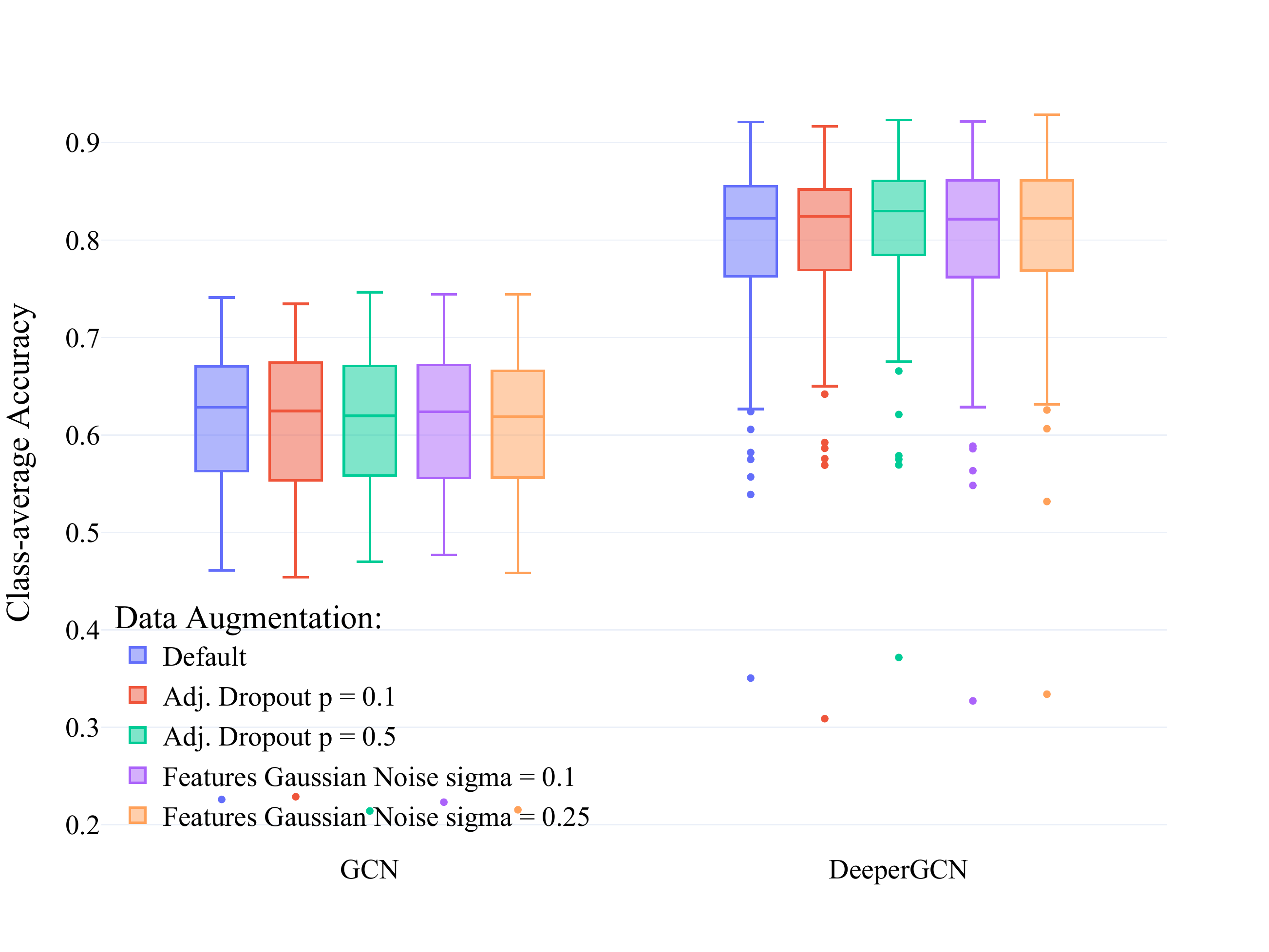}
  \caption{Comparison of the class-average accuracy for various combinations of network architecture, data augmentation technique, and parameters. The impact of data augmentation is negligible in all our experiments. Nevertheless, adjacency dropout (edge drop probability = 0.5) consistently improves accuracy for the DeeperGCN model.}
  \label{fig:data_aug}
\end{figure}

Data augmentation is commonly used in machine learning to avoid overfitting in a small dataset and improve generalization. We tested the impact of two simple approaches on our CellTypeGraph Benchmark. The first augmentation is to add Gaussian distributed noise to the node features, while the second approach is to use random dropout of edges in the cell adjacency graph, results are reported in \cref{fig:data_aug}. Our preliminary results show a small effect of data augmentation on the metrics, but more exhaustive experimentation is necessary to draw more solid conclusions.

\section{Baseline implementation details} 
In our baseline, we benchmark eight different graph neural network architectures. We report here the most salient implementation details. 
For \textit{GCN} \cite{kipf2016semi}, \textit{GraphSAGE}, \cite{hamilton2017inductive}, \textit{GIN} \cite{xu2018powerful}, \textit{GCNII} \cite{chen2020simple}, and \textit{DeeperGCN} \cite{li2019deepgcns, li2020deepergcn}; we followed the 
\textit{pytorch\_geometric} implementation \cite{Fey2019}. In all the aforementioned architectures the convolutions block are composed as follows: Graph Convolution $\rightarrow$ Relu activation \cite{agarap2018deep} $\rightarrow$ normalization $\rightarrow$ Dropout \cite{srivastava2014dropout}. The number of convolutions blocks used is an hyper parameter. In addition in \textit{DeeperGCN} and \textit{GCNII} and addition linear layer is added before the first graph convolution layer and after the last graph convolution layer. While for the remaining architectures \textit{GAT} \cite{velickovic2018graph}, \textit{GATv2} \cite{brody2021attentive}, and  \textit{TransformerGCN} \cite{vaswani2017attention, shi2020masked}, we used graph convolutions as implemented in \cite{Fey2019} but with some minor differences in the convolution block layout: Graph Convolution $\rightarrow$ normalization $\rightarrow$ Relu activation \cite{agarap2018deep} $\rightarrow$ Dropout \cite{srivastava2014dropout}. All source implementation are released at \url{https://github.com/hci-unihd/plant-celltype/blob/main/plantcelltype/graphnn/graph_models.py}.

\end{document}